\documentclass[runningheads]{llncs}

\usepackage{eccv}

\usepackage{eccvabbrv}

\usepackage{graphicx}
\usepackage{booktabs}
\usepackage{array}
\usepackage{multirow}
\usepackage{subcaption}
\usepackage{contour}
\usepackage{xcolor}
\usepackage{fontawesome5}
\usepackage{pgfplots}
\pgfplotsset{compat=1.16}
\usepgfplotslibrary{groupplots}

\definecolor{ftblue}{RGB}{0,83,159} 
\definecolor{layoutgreen}{RGB}{0,120,80}
\definecolor{correctionmissing}{HTML}{F58231}
\definecolor{correctionextra}{HTML}{4363D8}
\definecolor{correctionmodified}{HTML}{800000}
\definecolor{correctioncorrect}{HTML}{000000}
\definecolor{manualgray}{gray}{0.35}

\contourlength{0.15pt}

\usepackage{tikz}
\usetikzlibrary{positioning, arrows.meta, shapes.geometric, calc}

\newlength{\panelheight}
\newsavebox{\coverimagebox}
\newlength{\coverimagewidth}
\newlength{\coverimageheight}
\newlength{\flowchartimagewidth}
\newsavebox{\flowchartoriginalbox}
\newsavebox{\flowchartheatmapbox}
\newsavebox{\flowchartgraphbox}

\makeatletter
\NewDocumentCommand{\coverimage}{O{0pt} O{1} m m}{%
    \sbox{\coverimagebox}{\includegraphics{#4}}%
    \setlength{\coverimagewidth}{\wd\coverimagebox}%
    \setlength{\coverimageheight}{\ht\coverimagebox}%

    \pgfmathsetmacro{\sourceaspect}{%
        \strip@pt\coverimagewidth / \strip@pt\coverimageheight
    }%
    \pgfmathsetmacro{\targetaspect}{%
        \strip@pt#3 / \strip@pt\panelheight
    }%
    \pgfmathsetlengthmacro{\coverimagescaledshift}{#2*(#1)}%

    \begin{tikzpicture}
        \path[use as bounding box]
            (0,0) rectangle (#3,\panelheight);

        \clip
            (0,0) rectangle (#3,\panelheight);

        \ifdim\sourceaspect pt>\targetaspect pt
            \node[inner sep=0, yshift=\coverimagescaledshift] at
                (current bounding box.center)
                {\scalebox{#2}{%
                    \includegraphics[height=\panelheight]{#4}%
                }};
        \else
            \node[inner sep=0, yshift=\coverimagescaledshift] at
                (current bounding box.center)
                {\scalebox{#2}{%
                    \includegraphics[width=#3]{#4}%
                }};
        \fi
    \end{tikzpicture}%
}
\makeatother

\usepackage{listings}
\usepackage[accsupp]{axessibility}  % Improves PDF readability for those with disabilities.

\usepackage{hyperref}

\usepackage{orcidlink}

\begin{document}

% ---------------------------------------------------------------
% TODO REVIEW: Replace with your title
\title{Impact of Iterative Fine-Tuning on Transcription Accuracy in Complex Historical Sanskrit Manuscripts} 

% TODO REVIEW: If the paper title is too long for the running head, you can set
% an abbreviated paper title here. If not, comment out.
\titlerunning{Iterative Fine-Tuning of Layout Analysis and OCR}

% TODO FINAL: Replace with your author list. 
% Include the authors' OCRID for the camera-ready version, if at all possible.
% \author{First Author\inst{1}\orcidlink{0009-0001-1451-6759} \and
% Second Author\inst{2,3}\orcidlink{0000-0002-7980-6183} \and
% Third Author\inst{3}\orcidlink{2222--3333-4444-5555}}

% TODO FINAL: Replace with an abbreviated list of authors.
% \authorrunning{F.~Author et al.}
% First names are abbreviated in the running head.
% If there are more than two authors, 'et al.' is used.

% TODO FINAL: Replace with your institution list.
% \institute{Princeton University, Princeton NJ 08544, USA \and
% Springer Heidelberg, Tiergartenstr.~17, 69121 Heidelberg, Germany
% \email{lncs@springer.com}\\
% \url{http://www.springer.com/gp/computer-science/lncs} \and
% ABC Institute, Rupert-Karls-University Heidelberg, Heidelberg, Germany\\
% \email{\{abc,lncs\}@uni-heidelberg.de}}

\author{Kartik Chincholikar\orcidlink{0009-0001-1451-6759} \and
Kaushik Gopalan\orcidlink{0000-0002-7980-6183} \and
Mihir Hasabnis}

\authorrunning{K. Chincholikar et al.}

\institute{
Centre for Inter-disciplinary Artificial Intelligence (CAI),\\
FLAME University, Pune, India\\
\email{\{kartik.chincholikar,kaushik.gopalan\}@flame.edu.in}\\
\email{mihir.hasabnis@flame.edu.in}
}

\maketitle

\begin{abstract}
    Digitizing the text from handwritten historical manuscripts is required to make them easily accessible, preservable, and to enable historical scholars to study them in new ways. Historical manuscripts, however, often exhibit complex heterogeneous layouts and non-standard appearance due to period-specific writing styles, page textures, camera noise, and other nuisance factors, making them difficult to perform OCR on. To tackle this challenge, we introduce a local traditional OCR pipeline, which can be iteratively fine-tuned on the target manuscript at the layout-level and the appearance-level. By adapting to the target manuscript distribution, the proposed Traditional OCR pipeline makes better predictions on subsequent pages, causing iterative reduction in human annotation effort, which is expensive and time-consuming as it requires historical domain expertise. Using this pipeline, we digitize text from three complex historical Sanskrit manuscripts and introduce a dataset with granular layout-level annotations, along with Unicode annotations in the standard PAGE-XML format. We demonstrate quantitative gains due to iterative fine-tuning of the proposed traditional OCR pipeline, and also benchmark the performance of leading Multi-Modal Large Language Models on the introduced Dataset. Code and dataset are available at: \url{https://github.com/flame-cai/gnn-synthetic-layout-historical/}.
\end{abstract}

% TODO
% Code\footnote{\url{https://anonymous.4open.science/r/gnn-synthetic-layout-historical-2AEF/}}. Dataset: Coming Soon.
% update literature review
% check anonymous github
% Manuscript

\section{Introduction}
Digitization of historical manuscripts in Unicode Format makes them easily accessible to researchers and scholars in a digital format, avoiding the risk of damaging the original manuscripts, which are often in a fragile condition.
Such digitization also allows historical scholars to search through the manuscripts more quickly, study changes in word usage over time, and track the frequency with which certain ideas appear. % The National Manuscript Survey, India, 2026 further estimated 11 million manuscripts present in various repositories across the country~\cite{pib2026gyanbharatam}.

Digitization efforts of historical manuscripts often require specialized historical language and script expertise and are thus expensive and time-consuming to collect, causing the scarcity of annotated data. Furthermore, historical manuscript pages differ from modern digital or printed documents at the \textbf{Layout-level} and at the \textbf{Appearance-level}. Layout-level distribution shifts arise from high heterogeneity in historical page layouts—including marginalia, interlinear glosses, footnotes, and irregular, curved text-lines. Appearance-level distribution shifts occur due to variation in scribal styles, period-specific writing conventions, physical degradation artifacts (e.g., ink bleed-through, fading, staining) and nuisance factors (e.g., camera noise, image compression, paper or palm-leaf textures, uneven illumination, darkened scans).

\begin{figure}[!t]
    \centering

    % ---------------- Subfigure 1 ----------------
    \begin{subfigure}[t]{\textwidth}
        \centering
        \coverimage{\linewidth}{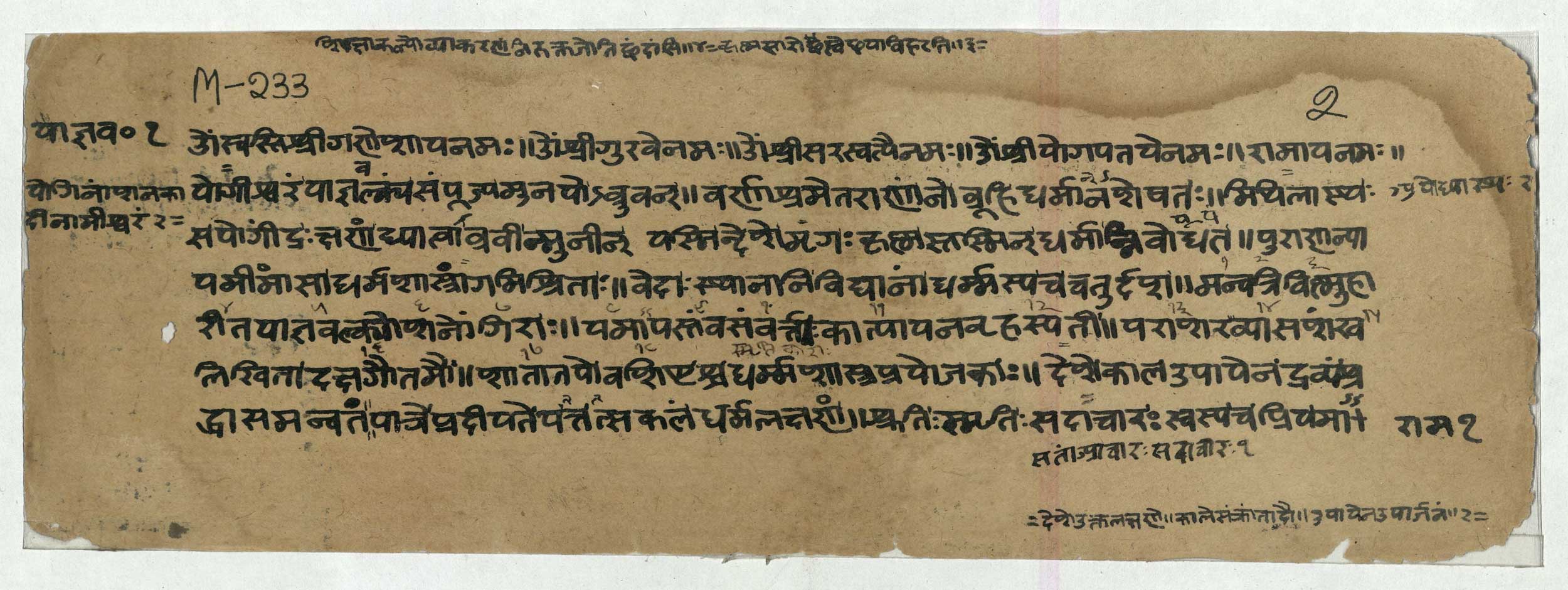}
        \caption{\textbf{Moderate Layout Manuscript:} \textit{Yajnavalakyasmritih (Acharadhyayah)}\cite{Yajna}}
        \label{fig:method-a}
    \end{subfigure}

    \vspace{0.5em}

    % ---------------- Subfigure 2 ----------------
    \begin{subfigure}[t]{\textwidth}
        \centering
        \coverimage{\linewidth}{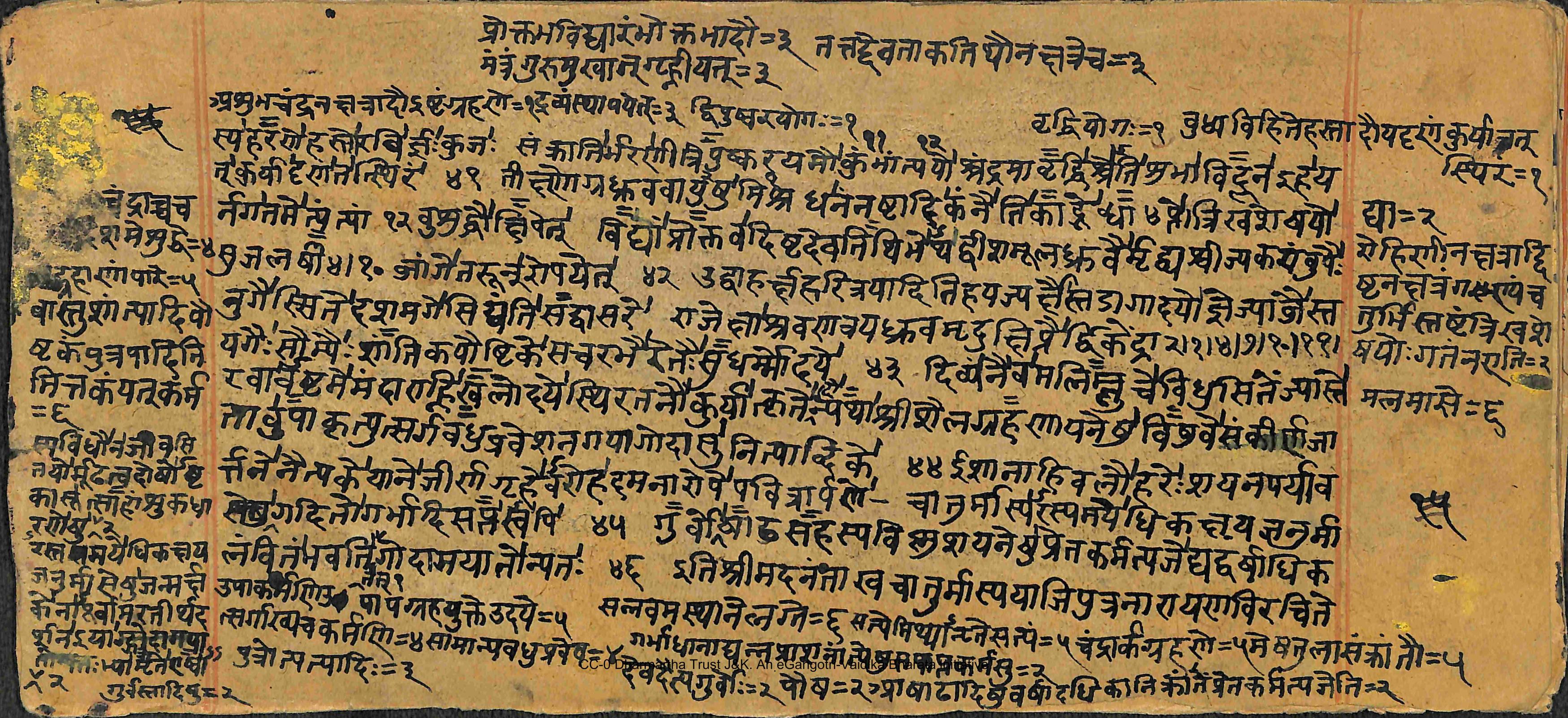}
        \caption{\textbf{Dense Layout Manuscript:} \textit{Muhurta Martanda}\cite{Muhurta}}
        \label{fig:method-b}
    \end{subfigure}

    \vspace{0.5em}

    % ---------------- Subfigure 3 ----------------
    \begin{subfigure}[t]{\textwidth}
        \centering
        \coverimage[30mm]{\linewidth}{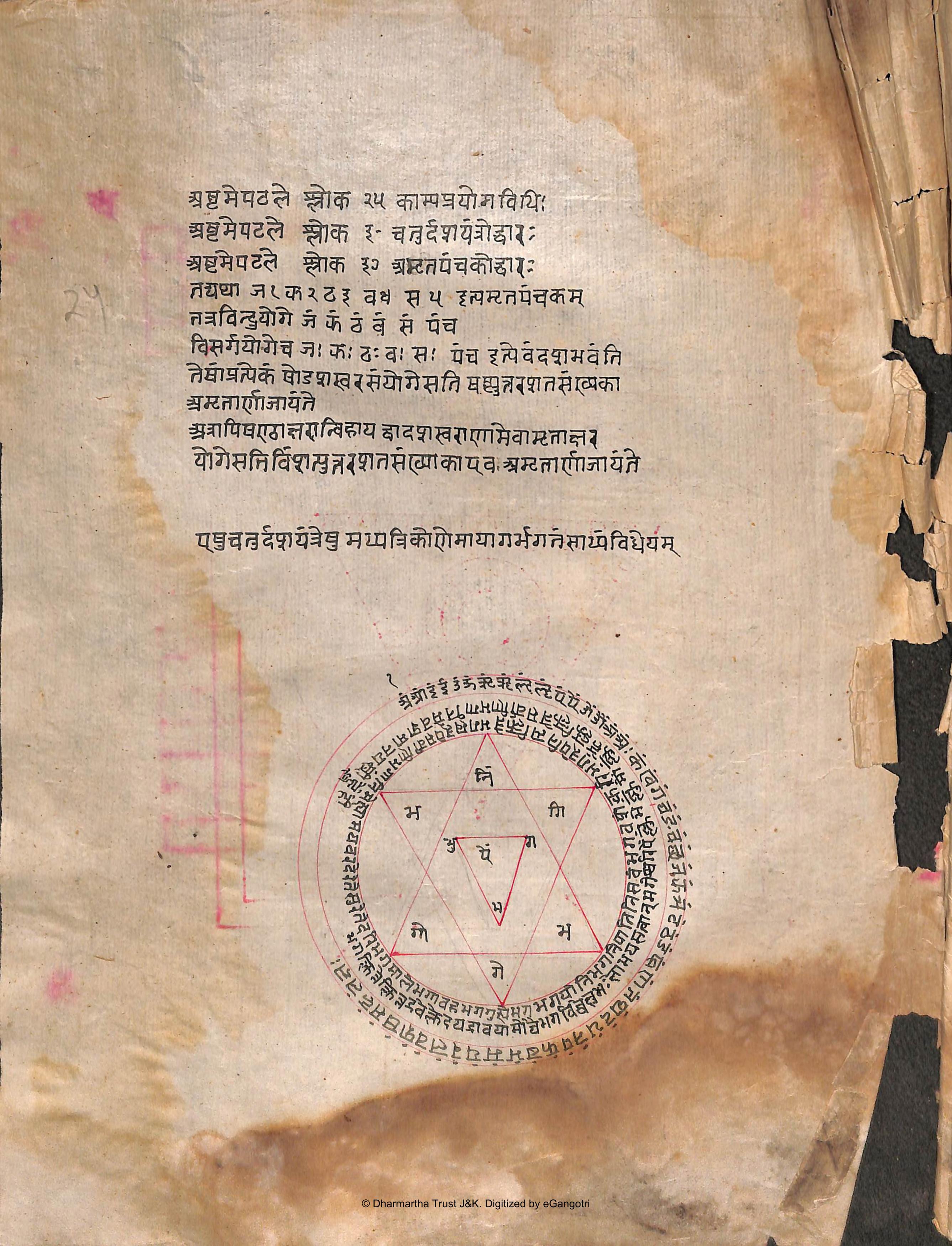}
        \caption{\textbf{Circular Layout Manuscript:}\textit{Tantra Raj With Yantra And Mantra Uddhara}\cite{Tantra}}
        \label{fig:method-c}
    \end{subfigure}

    \caption{
        \textbf{The manuscripts.}\quad Samples of pages from the three manuscripts digitized in this case study. The manuscripts exhibit \textit{non-standard layout variations}, including dense marginalia, circular text, interlinear commentary, irregular text-line orientation, as well as \textit{non-standard appearance variations} such as ink bleed-through, faded ink, paper degradation, scribe-specific handwriting styles, and period-specific orthographic conventions.
    }
    \label{fig:dataset_illustration}
\end{figure}

In this work, we thus digitize three historical Sanskrit manuscripts with complex layouts and non-standard appearances, as illustrated in~\cref{fig:dataset_illustration}, using a traditional two-step OCR pipeline: first, we perform layout analysis and segment individual text-line images from manuscript pages, and second, we transcribe those text-line images into machine-readable Unicode text. The proposed traditional pipeline allows us to iteratively fine-tune and adapt the pipeline to the target manuscript - at the layout-level and at the appearance-level, thus progressively reducing the burden of human annotation on subsequent pages. This is important, as historical data is scarce, time-consuming, and expensive to annotate.
\medskip
\noindent With this context, our work makes two primary contributions:

\noindent \textbf{1) Fine-tunable Traditional OCR pipeline.}\quad We introduce an open-source digitization pipeline, which can be iteratively fine-tuned from human supervision at the layout-level and the appearance-level, allowing progressive reduction in the burden of human annotation of subsequent pages.

\noindent \textbf{2) Curated Dataset.}\quad We introduce a high-quality dataset, which is richly annotated at the layout-level and the appearance-level, and is available in the standard PAGE-XML format~\cite{pletschacher2010page}~\footnote{\url{https://github.com/PRImA-Research-Lab/PAGE-XML}}.

\section{Literature Review}
Modern Multi-Modal LLMs have made significant progress in their OCR capabilities~\cite{yin2026unlimitedocrworks, wei2025deepseek, cui2025paddleocr, poznanski2025olmocr, liao2023doctr, xu2020layoutlm}. However, the benchmark datasets~\cite{ouyang2025omnidocbenchbenchmarkingdiversepdf, 10.1007/978-3-032-04614-7_2,Venna_2026_CVPR} mostly consider documents in a printed format or in a standard digital format. Therefore, at the date of writing, performing OCR on historical documents with complex, dense layouts with non-standard appearance using Modern Multi-Modal LLMs remains inconsistent~\cite{crosilla2025benchmarking,he2025seeing,coquenet2023dan}, because of a distribution shift of the target distribution from the pre-training distribution, and due to scarcity of annotated data from the target distribution.

To efficiently learn from scarce data, traditional OCR pipelines are hence employed to digitize historical manuscripts~\cite{kiessling2019kraken,kahle2017transkribus,rohit2017,saluja2017framework,adiga2018improving,das2021enhancing,vinitha2016error,indsenz,surya,tesseract,paddleocr_0000,liwicki3,papadopoulos2013impact,trivedi2019hindola}. Traditional pipelines broadly consist of two steps: (a) Text-line Segmentation, where the layout of the page is analyzed and individual text-line images are segmented from manuscript pages and (b) text recognition OCR, where the text-line images are transcribed into machine-readable Unicode text. 

Early text-line segmentation methods follow a projection profile approach successfully; however, they are limited to applications where the manuscript layouts are known a-priori~\cite{chamchong_text_2012, nguyen2022effective, chincholikar2025case}. More recent methods perform end-to-end text-line segmentation (sometimes with subsequent algorithmic post-processing) by predicting bounding polygons or by predicting pixel level masks~\cite{sharan2021palmira,vadlamudi2023seamformer,jindal2023text,fizaine2024historical,boillet2021multiple,vadlamudi2023seamformer,kiessling_modular_2020,oliveira2018dhsegment,gruning2018read, Boillet_2021}. Performing such dense end-to-end pixel-level predictions is, however, not robust to distribution shifts commonly encountered in historical manuscripts~\cite {das2021enhancing,aubreville2021quantifying,agrawal2025linetr, hendrycks2021natural,chincholikar2026towards}. To alleviate this, instead of performing dense end-to-end pixel-level prediction, recent methods LineTR~\cite{vaibav2024linetr} and CurT~\cite{kiessling2022curt} use deep learning to extract information from the manuscript images and use it to predict \textit{the parameters defining the geometry} of piecewise line segments or cubic Bézier curves representing the text lines, thus making effective use of inductive priors and the geometric structure of text-lines. To handle complex unconstrained historical layouts, systems like Kodym and Hradi\v{s}~\cite{kodym2021pagelayoutanalysisunconstrained} jointly estimate baselines, line heights, and pixel-wise text orientations to extract clean line regions. Challenges in text-line segmentation motivated the FEST Competition 2025 (Few-Shot Text-Line Segmentation)~\cite{zottin2025icdar}, with the aim of stimulating research in the design of methods that generalize to target manuscripts when trained on scarce data. 

Once the text-line images are segmented, the text content from the text-line images can be recognized in a Unicode format using methods which use a Convolutional Neural Network(CNN) for extracting images features, followed by a BiLSTM or an RNN for sequential modelling, with a CTC loss and decoder~\cite{graves2007multi,sankaran2013devanagari,karayil2015segmentation,shi2016end,dwivedi2020ocr}. More recent methods use the Transformer architecture~\cite{vaswani2017attention}, where an image Transformer extracts the visual features
and a text Transformer performs the language modeling and decoding~\cite{li2022trocrtransformerbasedopticalcharacter}. Beyond architectural designs, recent line recognition methods address domain-specific challenges in historical HTR: AT-ST~\cite{Ki__2021} introduces self-training adaptation to train line recognizers when target domain transcripts are scarce, while TS-Net~\cite{Koh_t_2021} enables a single recognizer to switch dynamically between different transcription styles (e.g., diplomatic versus modernized output).

In traditional pipelines, layout analysis and text-line segmentation are thus prerequisites for performing line level text recognition OCR, and errors in layout analysis and text-line segmentation can negatively impact the downstream text recognition OCR task. Hence, manual work and human supervision might still be required at the layout analysis stage~\cite{a16030136}. For manuscripts with non-standard layouts and unconventional reading order, character segmentation (instead of text-line segmentation) is a promising task decomposition~\cite{sharma2026episamcharactersegmentationchallenging,clanuwat2019kuronet}, offering increased flexibility to perform layout analysis. 

% CITE VISUAL QUESTION ANSWERING, CITE LAYOUT DATASET, UNILIPI
% CITE TEXT-LINE SEGMENTATION ONLY METHODS
% CITE TRADITIONAL PIPELINE METHODS
% CITE VLM METHODS
% CITE Printed VLM methods
% CITE OTHER DATASETS
% CITE FOR PRINTED
% \subsection{Benchmark 2\cite{Venna_2026_CVPR}}
% \subsubsection{Structure}:
% Data Generation Pipeline
% State what the inputs are, and what the outputs could be..
% \subsection{Phrases}:
% SEGMENT FIRST, ANSWER LATER vs ANSWER FIRST, SEGMENT LATER!!
% bleed-suppression loss
% evidence localization

\section{Dataset}

\begin{table}[t]
\centering
\caption{Dataset statistics}
\label{tab:dataset-content-statistics}
\small
\setlength{\tabcolsep}{4pt}
\renewcommand{\arraystretch}{1.12}
\begin{tabular}{@{}lcrrrrrr@{}}
\toprule
Manuscript & Pages & \multicolumn{3}{c}{\shortstack{Text-lines\\per page}} & \multicolumn{3}{c}{\shortstack{Grapheme clusters\\per page}} \\
\cmidrule(lr){3-5} \cmidrule(lr){6-8}
 & & \multicolumn{1}{c}{Min.} & \multicolumn{1}{c}{Max.} & \multicolumn{1}{c}{Mean} & \multicolumn{1}{c}{Min.} & \multicolumn{1}{c}{Max.} & \multicolumn{1}{c}{Mean} \\
\midrule
Moderate Layout & 15 & 12 & 28 & 21.07 & 285 & 525 & 392.27 \\
Dense Layout & 7 & 15 & 78 & 44.43 & 285 & 1098 & 691.86 \\
Circular Layout & 9 & 10 & 65 & 28.33 & 164 & 434 & 258.89 \\
\bottomrule
\end{tabular}
\end{table}

The dataset consists of three historical Sanskrit manuscripts: \textit{Yajnavalakyasmritih (Acharadhyayah)}, \textit{Muhurta Martanda}, and \textit{Tantra Raj With Yantra And Mantra Uddhara}. We will henceforth refer to the manuscripts as \textit{Moderate Layout Manuscript}, \textit{Dense Layout Manuscript}, and \textit{Circular Layout Manuscript} respectively, based on their page layouts as illustrated in~\cref{fig:dataset_illustration}.~\cref{tab:dataset-content-statistics} shows the number of pages, the number of text-lines, and grapheme clusters~\cite {unicode-uax29-2025,dwivedi-gopalan-2026-comparative} in each manuscript. A grapheme cluster corresponds to a visually identifiable unit in the script, but it is made up of two or more Unicode points.

\begin{figure*}[t]
  \centering
  \begin{subfigure}[t]{0.245\textwidth}
    \includegraphics[width=\linewidth]{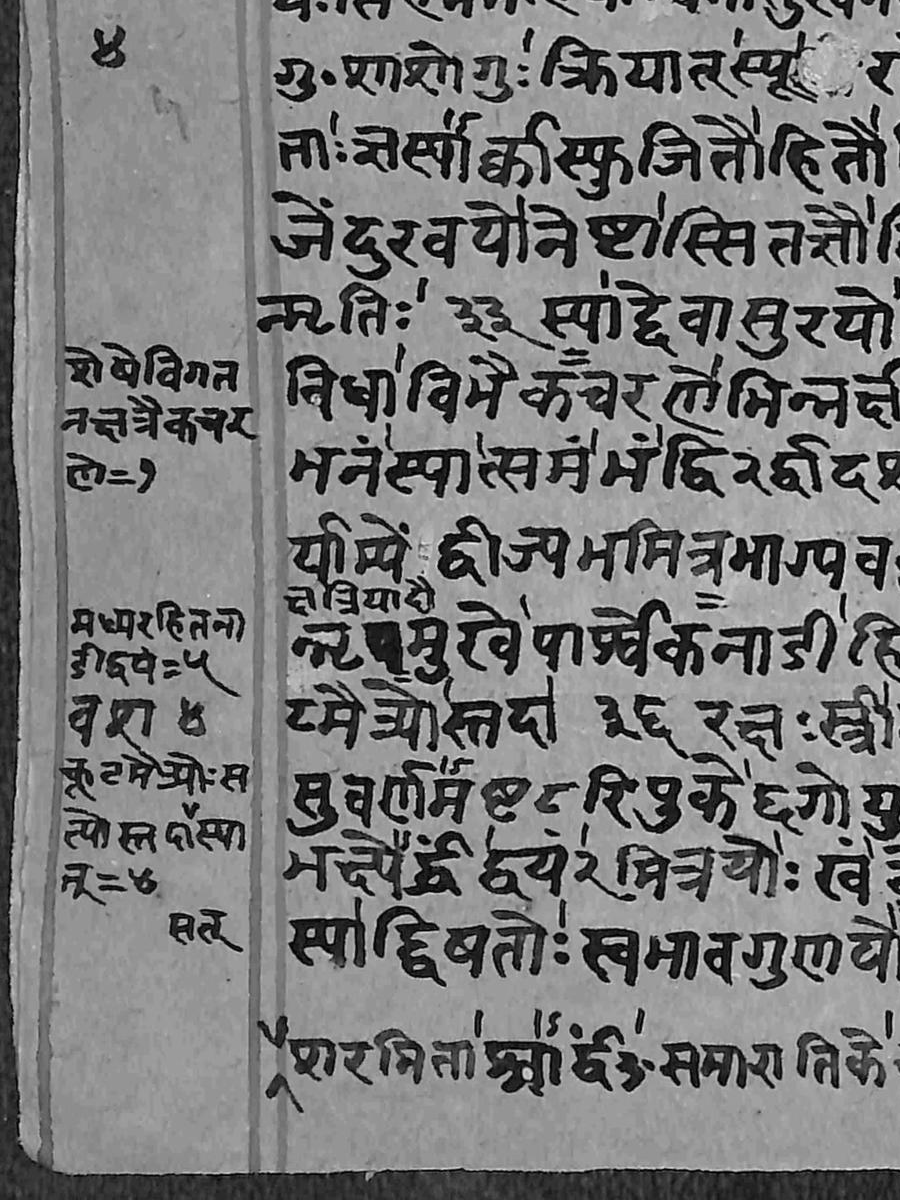}
    \caption{Original image}
  \end{subfigure}\hfill
  \begin{subfigure}[t]{0.245\textwidth}
    \includegraphics[width=\linewidth]{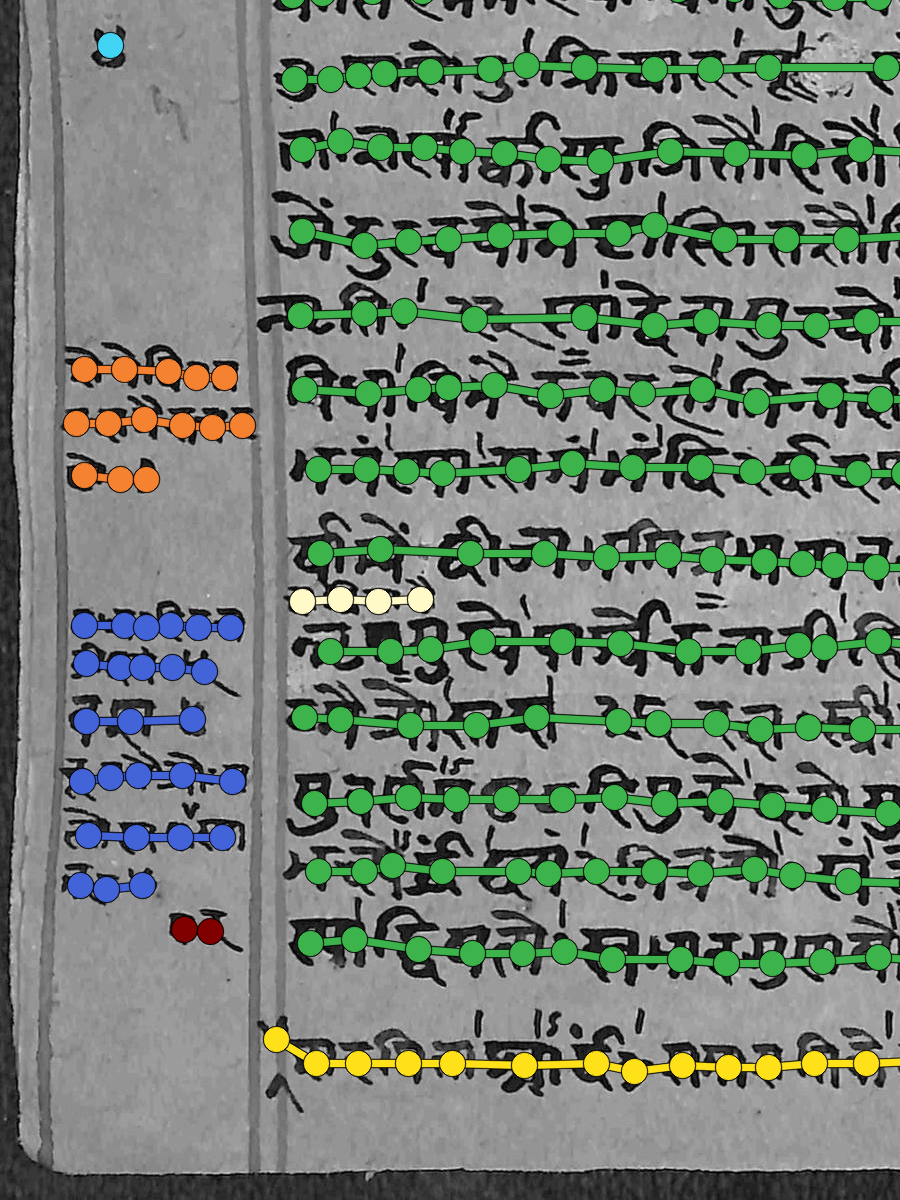}
    \caption{Text-region annotation}
  \end{subfigure}\hfill
  \begin{subfigure}[t]{0.245\textwidth}
    \includegraphics[width=\linewidth]{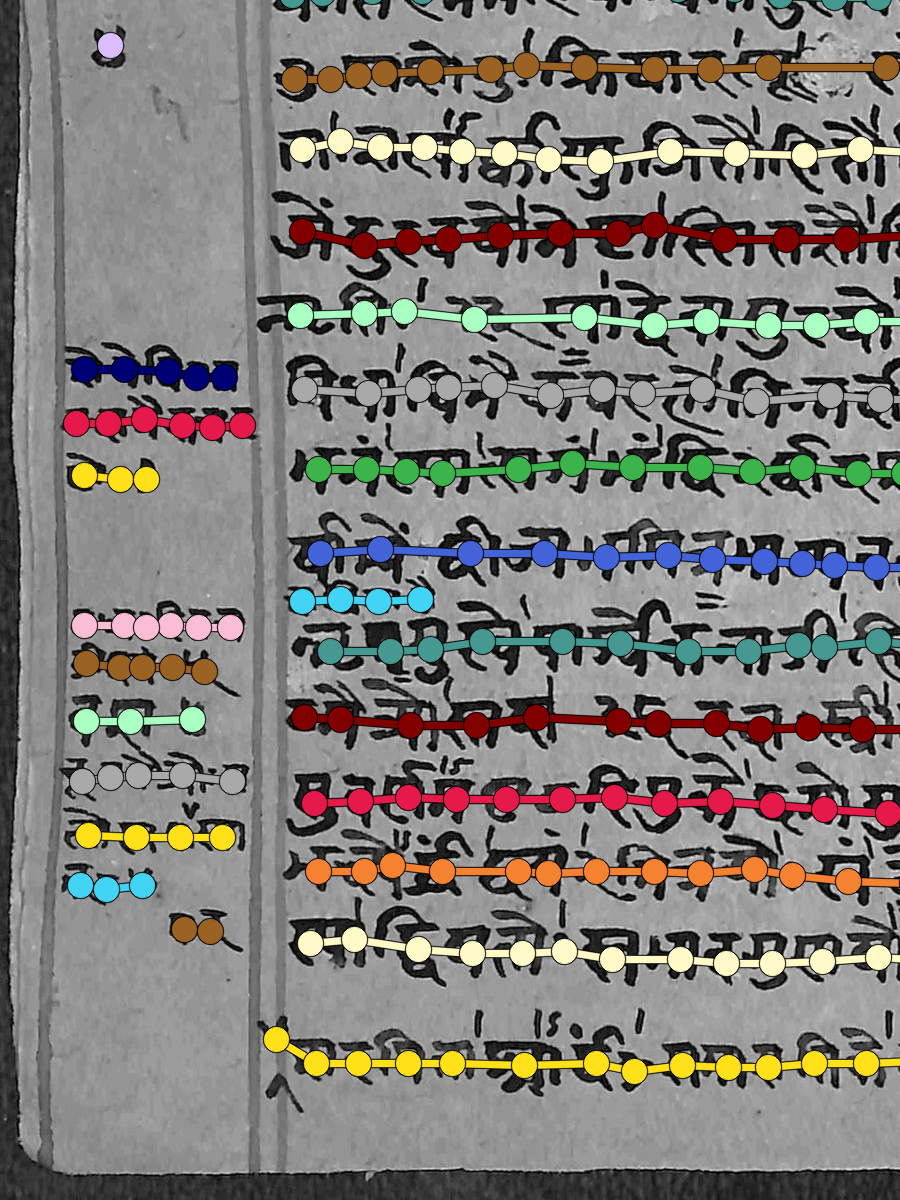}
    \caption{Text-line annotation}
  \end{subfigure}\hfill
  \begin{subfigure}[t]{0.245\textwidth}
    \includegraphics[width=\linewidth]{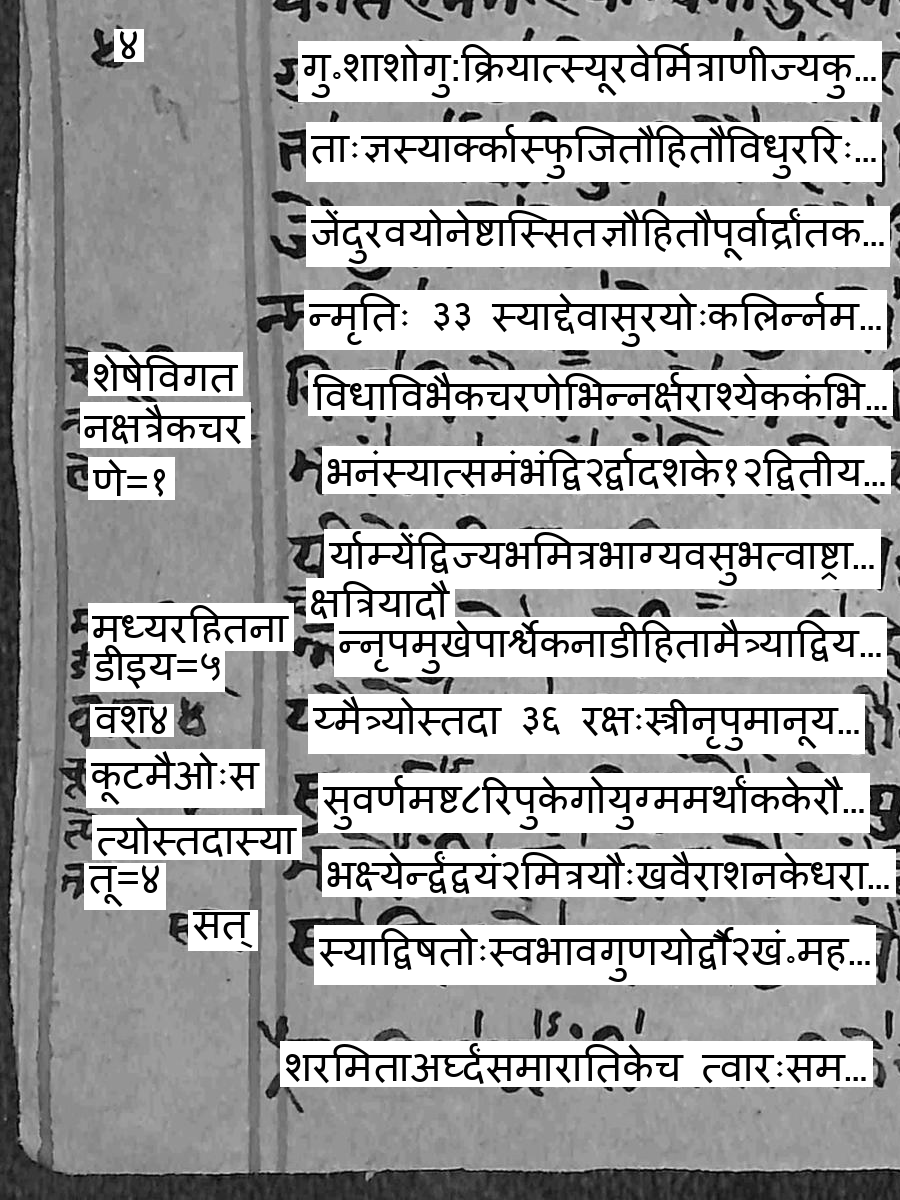}
    \caption{Unicode annotation}
  \end{subfigure}
  \caption{Rich granular annotations at the Layout-level and Appearance-level. The dataset is also exported in the standard PAGE-XML format.}
  \label{fig:page-10-annotation-views}
\end{figure*}

\medskip
\noindent\textbf{Annotation Methodology.}\quad As illustrated in Figure~\cref{fig:page-10-annotation-views}, we annotate the manuscripts at the Layout-level and Appearance-level. For layout-level annotations, we consider each character (or grapheme cluster) of the manuscript page as a node, with edges connecting nodes with their neighbours in the text-line together. Thus, all nodes belonging to the same text-line have the same label, as shown in~\cref{fig:page-10-annotation-views}(c). Similarly, all nodes belonging to the same text-region have the same label, as shown in~\cref{fig:page-10-annotation-views}(b). The user can hover over the predicted graph while pressing and holding keys "a" or "d" to add or delete edges, respectively. A "right-click" or "left-click" adds or deletes nodes, respectively. Similarly, text-regions can be annotated by pressing and holding "e" and hovering over the text-lines in the text-region. Each graph-based text-line is explicitly linked to it's corresponding Unicode text content, as illustrated in~\cref{fig:page-10-annotation-views}(d).

While annotating the dataset, we made the following assumptions: (a) We annotate text-regions, such that the reading order of the text-lines inside the text-region must be unambiguous. (b) Given the nature of the marginalia and commentary, the reading order of the text-regions is ambiguous. (c) The reference annotation symbols and numbers that link the main text to the commentary are not annotated. (d) When a watermark overlaps with the handwritten text-content, we give precedence to the handwritten text-content. (e) If the text-line segmentation incorrectly excludes a diacritic mark in the segmentation, we still annotate it in the Ground-Truth Unicode annotation. (f) If a character or a grapheme cluster is scratched out, causing the character to be illegible, we do not annotate it.

\medskip
\noindent\textbf{Selection Criteria.}\quad
The proposed Traditional Pipeline can be used to digitize rare historical manuscripts that fit the following selection criteria: (a) The frozen character segmentation model CRAFT~\cite{baek2019character} should perform satisfactorily, and (b) A pre-trained text recognition OCR model is available for the manuscript's script. We believe that these selection criteria should apply to most of the Sanskrit manuscript images archived in culture preservation projects like the eGangotri project\footnote{\url{https://egangotri.org/}}, and GyanBharatam\footnote{\url{https://gyanbharatam.com/}}.

\section{Method}

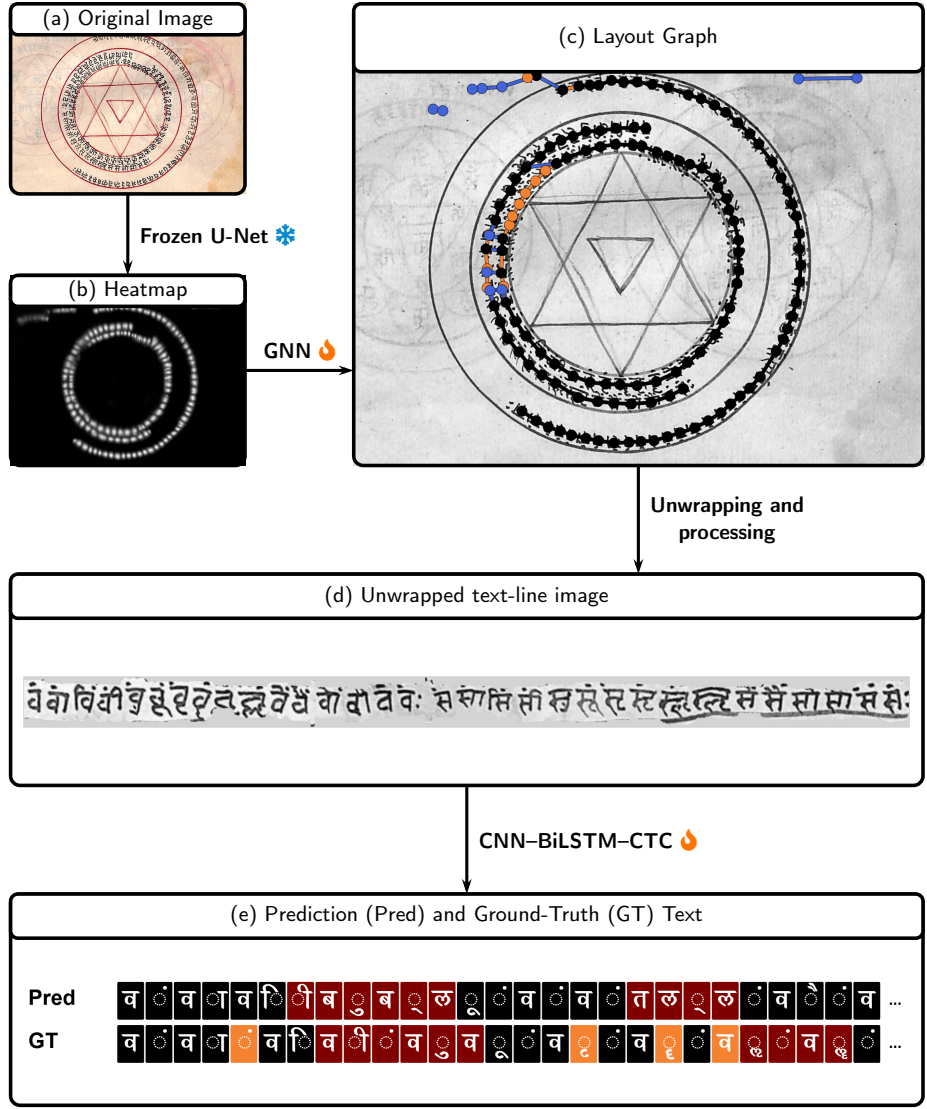
\begin{figure}[htbp]
\centering
\setlength{\panelheight}{2.80cm}
\setlength{\flowchartimagewidth}{3.45cm}
\def\imagecropVShift{-9mm}
\def\flowchartimagezoom{1.8}
\sbox{\flowchartoriginalbox}{%
    \coverimage[\imagecropVShift][\flowchartimagezoom]{\flowchartimagewidth}{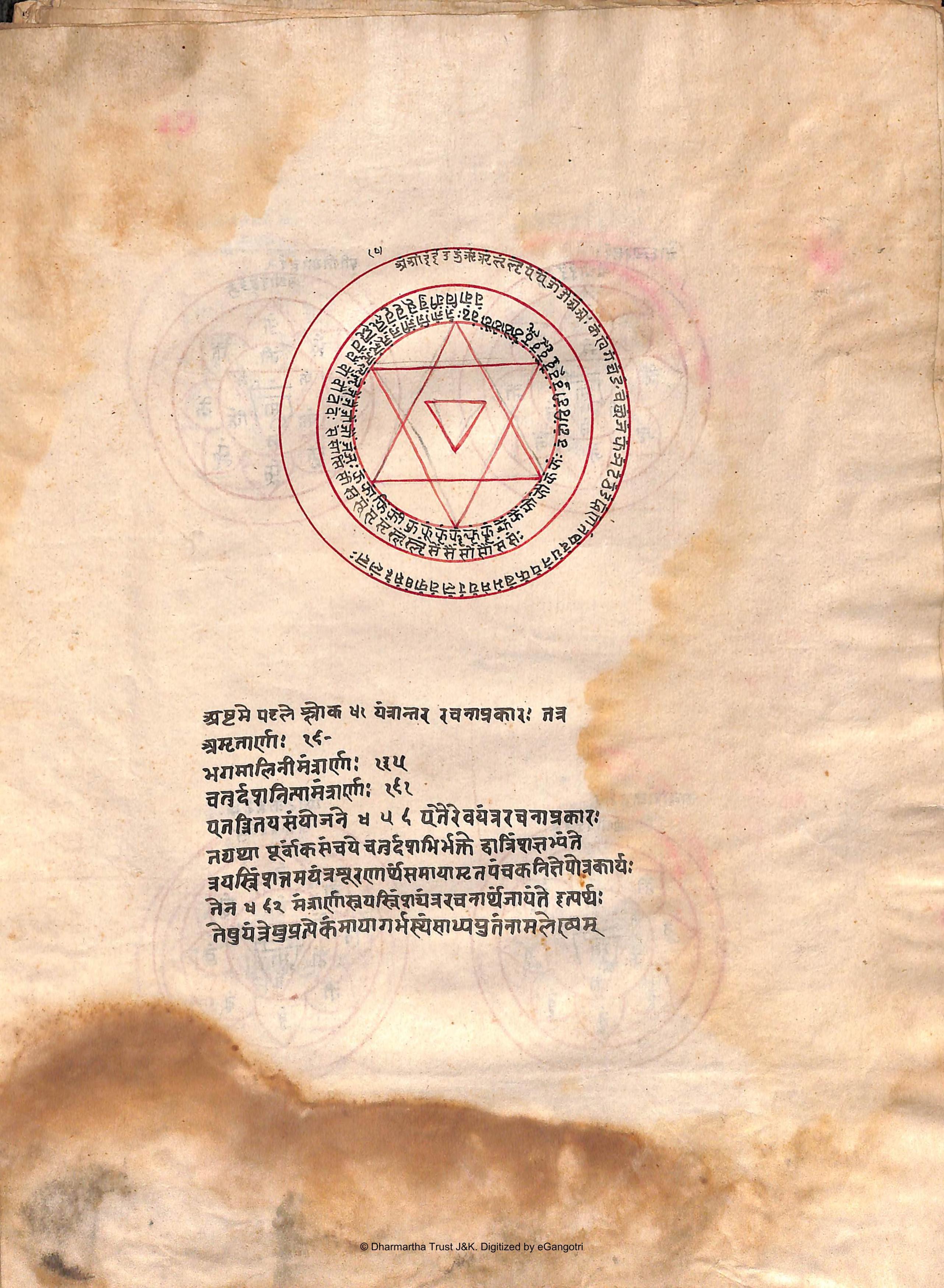}%
}
\sbox{\flowchartheatmapbox}{%
    \coverimage[\imagecropVShift][\flowchartimagezoom]{\flowchartimagewidth}{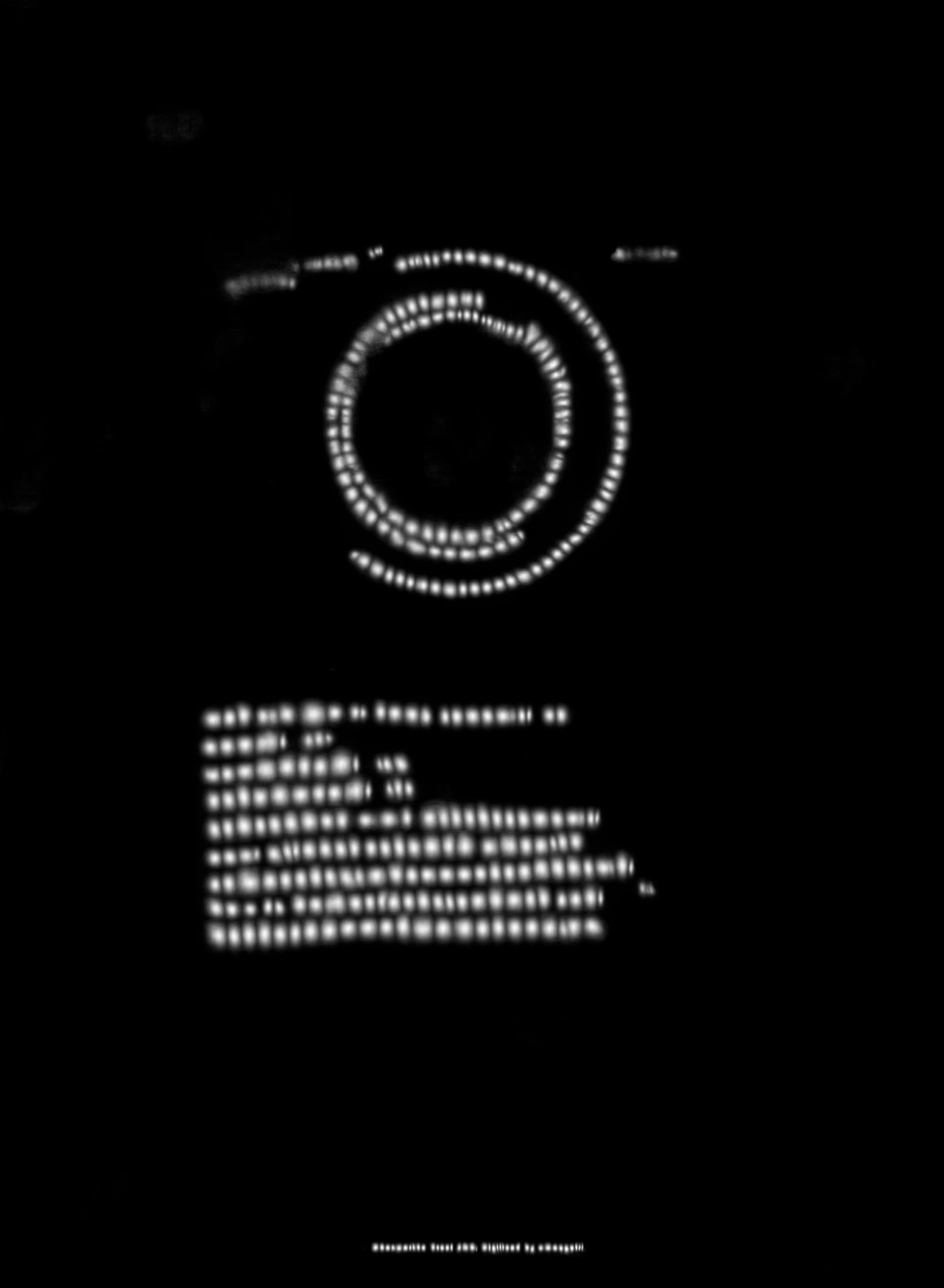}%
}
% Panel (c) spans both small panels plus their enlarged 1.20 cm gap.
% Its height, width, and crop offset use the same 6.80/2.80 scale factor
% so that it shows exactly the same manuscript crop as panels (a) and (b).
\setlength{\panelheight}{6.80cm}
\setlength{\flowchartimagewidth}{8.3786cm}
\def\graphimagecropVShift{-21.8571mm}
\sbox{\flowchartgraphbox}{%
    \coverimage[\graphimagecropVShift][\flowchartimagezoom]{\flowchartimagewidth}{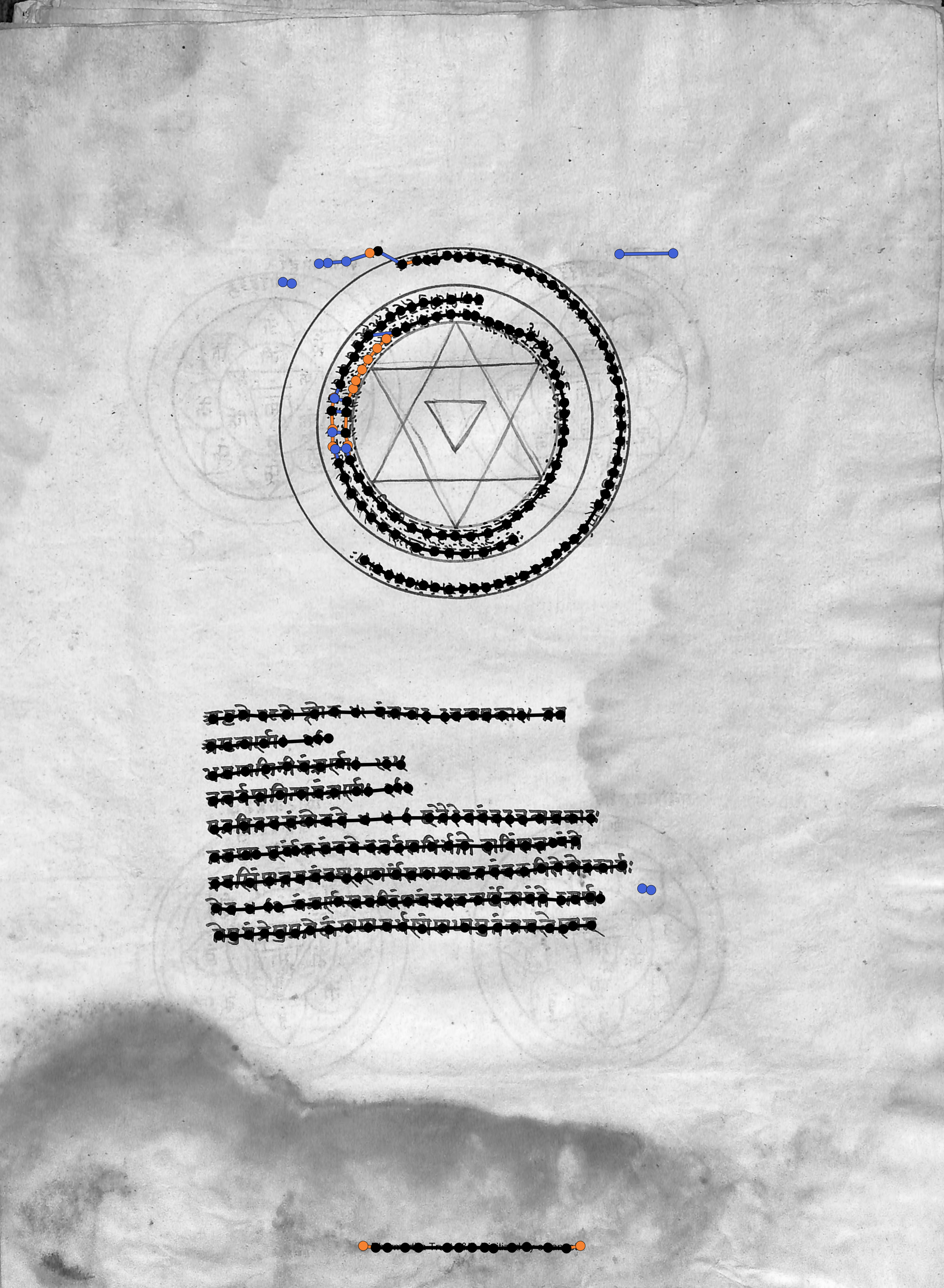}%
}
\resizebox{\columnwidth}{!}{%
\begin{tikzpicture}
% ------------------- TIKZ LIBRARIES -------------------
\usetikzlibrary{arrows.meta, calc}

% ------------------- CENTRALIZED CONTROL VARIABLES -------------------
\def\nodedistH{1.6cm}
\def\nodedistV{1.2cm}
\def\outputdistV{1.6cm}
\def\outputlabelheight{0.65cm}
\def\pipelineframelw{1.5pt}

% ------------------- STYLE DEFINITIONS -------------------
% --- Centralized color definitions for easy configuration ---
\colorlet{branch1color}{black}

% Base style for all frames
\tikzset{base frame/.style={
    draw=gray!80, thick, rounded corners=4pt, inner sep=4pt, fill=gray!10
}}

% Style for the framed pipeline images
\tikzset{branch 1 frame/.style={
    draw=branch1color,
    line width=\pipelineframelw
}}

\tikzset{
  arrow/.style={
    -{Stealth[length=2.2mm,width=1.4mm]}, % clean arrow-tip definition
    thick,
    rounded corners=2pt,
    line width=1.3pt,
    draw=branch1color
  }
}

% Style for the overlay labels inside frames
\tikzset{overlay label/.style={
    branch 1 frame,
    font=\small\sffamily,
    fill=white,
    fill opacity=1.0,
    text opacity=1,
    align=center,
    anchor=north,
    rounded corners=3pt,
    inner xsep=5pt,
    inner ysep=2pt,
    outer sep=0pt
}}
\tikzset{pipeline arrow label/.style={
    font=\small\sffamily\bfseries,
    fill=white,
    fill opacity=0.95,
    text opacity=1,
    inner xsep=2pt,
    inner ysep=1pt,
    outer sep=0pt
}}
% ------------------- NODE PLACEMENT -------------------
% --- Stack (a) and (b) in the left column ---
\node[inner sep=0pt, outer sep=0pt] (original_image)
    {\usebox{\flowchartoriginalbox}};
\node[inner sep=0pt, outer sep=0pt,
      below=\nodedistV of original_image] (heatmap)
    {\usebox{\flowchartheatmapbox}};

% --- Panel (c) spans the full height of the stacked left column ---
\node[inner sep=0pt, outer sep=0pt, anchor=north west] (heuristic)
    at ($(original_image.north east)+(\nodedistH,0)$)
    {\usebox{\flowchartgraphbox}};

% --- Draw the panel frames after the image content so the images stay beneath them ---
\foreach \panel in {original_image,heatmap,heuristic}{
    \draw[branch 1 frame, rounded corners=4pt]
        (\panel.north west) rectangle (\panel.south east);
}

% ------------------- PLACING OVERLAY LABELS -------------------
% --- Full-width labels centered on the top edge of each framed panel ---
\path let \p1 = (original_image.north west), \p2 = (original_image.north east) in
    node[overlay label, minimum height=0.45cm, text width=(\x2-\x1)-10pt]
    at (original_image.north) {(a) Original Image};
\path let \p1 = (heatmap.north west), \p2 = (heatmap.north east) in
    node[overlay label, minimum height=0.45cm, text width=(\x2-\x1)-10pt]
    at (heatmap.north) {(b) Heatmap};
\path let \p1 = (heuristic.north west), \p2 = (heuristic.north east) in
    node[overlay label, minimum height=1.00cm, text width=(\x2-\x1)-10pt]
    at (heuristic.north) {(c) Layout Graph };

% --- Full-width OCR outputs below panels (a)--(c) ---
% 13.4286 cm = 3.45 cm for (a)/(b) + 1.60 cm gap + 8.3786 cm for (c).
\node[inner sep=0pt, outer sep=0pt, anchor=north west] (unwrapped_text_line_image)
    at ($(original_image.west |- heuristic.south)
        +(0,-\outputdistV)+(0,-\outputlabelheight)$)
    {\includegraphics[
        width=13.4286cm,
    ]{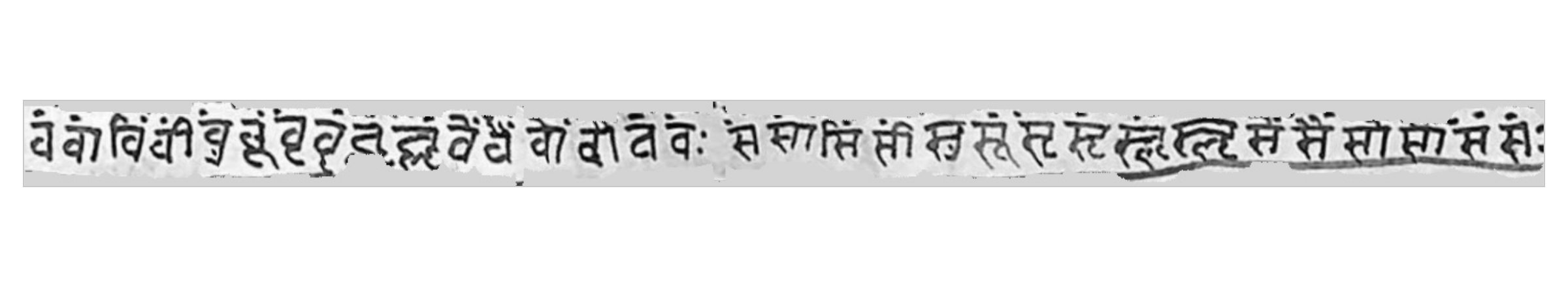}};
\coordinate (unwrapped_text_line_north_west) at
    ($(unwrapped_text_line_image.north west)+(0,\outputlabelheight)$);
\coordinate (unwrapped_text_line_north_east) at
    ($(unwrapped_text_line_image.north east)+(0,\outputlabelheight)$);
\coordinate (unwrapped_text_line_north) at
    ($(unwrapped_text_line_north_west)!0.5!(unwrapped_text_line_north_east)$);
\coordinate (unwrapped_text_line_entry) at
    (heuristic.south |- unwrapped_text_line_north);
\draw[branch 1 frame, rounded corners=4pt]
    (unwrapped_text_line_north_west) rectangle (unwrapped_text_line_image.south east);
\path let
    \p1 = (unwrapped_text_line_north_west),
    \p2 = (unwrapped_text_line_north_east)
in node[
    overlay label,
    minimum height=\outputlabelheight,
    text width=(\x2-\x1)-10pt
] at (unwrapped_text_line_north) {(d) Unwrapped text-line image};

\node[inner sep=0pt, outer sep=0pt, anchor=north west] (unicode_visualization_image)
    at ($(unwrapped_text_line_image.south west)
        +(0,-\outputdistV)+(0,-\outputlabelheight)$)
    {\includegraphics[
        width=13.4286cm
    ]{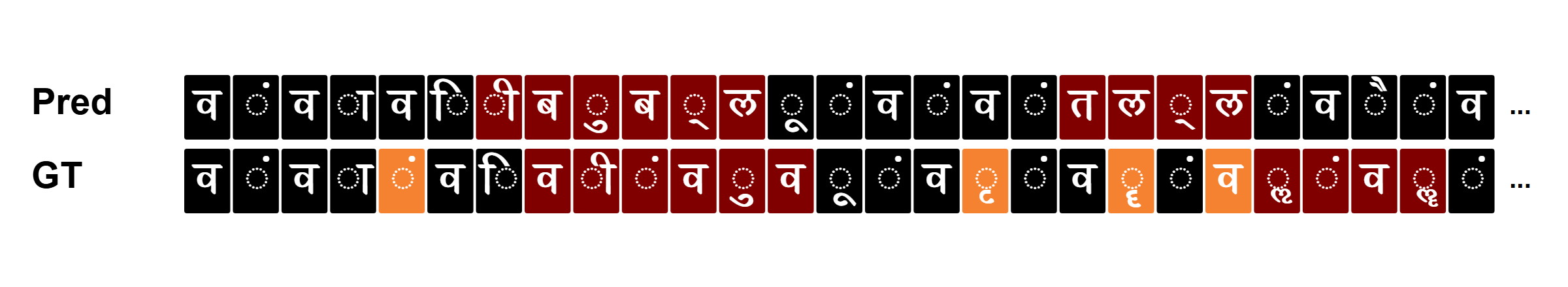}};
\coordinate (unicode_visualization_north_west) at
    ($(unicode_visualization_image.north west)+(0,\outputlabelheight)$);
\coordinate (unicode_visualization_north_east) at
    ($(unicode_visualization_image.north east)+(0,\outputlabelheight)$);
\coordinate (unicode_visualization_north) at
    ($(unicode_visualization_north_west)!0.5!(unicode_visualization_north_east)$);
\coordinate (unicode_visualization_entry) at
    (unwrapped_text_line_image.south |- unicode_visualization_north);
\draw[branch 1 frame, rounded corners=4pt]
    (unicode_visualization_north_west) rectangle (unicode_visualization_image.south east);
\path let
    \p1 = (unicode_visualization_north_west),
    \p2 = (unicode_visualization_north_east)
in node[
    overlay label,
    minimum height=\outputlabelheight,
    text width=(\x2-\x1)-10pt
] at (unicode_visualization_north) {(e) Prediction (Pred) and Ground-Truth (GT) Text };

% ------------------- DRAWING ARROWS -------------------
\draw[arrow] (original_image.south) --
    node[pipeline arrow label, midway, right=3pt]
        {Frozen U-Net~\textcolor{cyan!70!blue}{\faSnowflake}}
    (heatmap.north);
\draw[arrow] (heatmap.east) --
    node[pipeline arrow label, midway, above=3pt]
        {GNN~\textcolor{orange!90!red}{\faFire}}
    (heuristic.west |- heatmap.east);
\draw[arrow] (heuristic.south) --
    node[pipeline arrow label, midway, right=3pt]
        {\shortstack{Unwrapping and\\processing}}
    (unwrapped_text_line_entry);
\draw[arrow] (unwrapped_text_line_image.south) --
    node[pipeline arrow label, midway, right=3pt]
        {CNN--BiLSTM--CTC~\textcolor{orange!90!red}{\faFire}}
    (unicode_visualization_entry);

\end{tikzpicture}
}
\vspace{0mm}
\caption{\textbf{Iterative Fine-tuning Pipeline.}\quad The frozen~\textcolor{cyan!70!blue}{\faSnowflake} U-Net CRAFT~\cite{baek2019character} detects character locations in the original image (a) as a heatmap (b). Next, the GNN predicts the layout graph (c) where characters (or grapheme clusters) belonging to the same text-line are connected. In the next step, text-line images are prepared using the GNN predictions, and curved text (if any) is unwrapped and straightened (d). Finally, the CNN--BiLSTM--CTC takes in the text-line images as input, and predicts its text contents as Unicode text (e). In both (c) and (e), the colors denote differences between the predictions and the ground-truth, and hence illustrate the scope of improvement from iterative fine-tuning of the ~\textcolor{orange!90!red}{\faFire} GNN and ~\textcolor{orange!90!red}{\faFire} CNN--BiLSTM--CTC. \textcolor{correctionmissing}{\textbf{Orange}} denotes missing nodes, edges, or Unicode characters. \textcolor{correctionextra}{\textbf{Blue}} denotes extra nodes, edges, or Unicode characters. \textcolor{correctionmodified}{\textbf{Maroon}} denotes modified text, and \textcolor{correctioncorrect}{\textbf{Black}} denotes correct nodes, edges, and unicode text.
}
\label{fig:human_in_loop_pipeline}
\end{figure}

Historical manuscripts can vary at the \emph{Layout-level}, concerning where text is placed: dense pages, marginalia, interlinear text, and curved or circular text-lines, and at the \emph{Appearance-level}, concerning how text looks: handwriting style, period-specific writing conventions, paper texture, and uneven scans.~\cref{fig:human_in_loop_pipeline} shows the traditional pipeline, which allows fine-tuning at the Layout-level and at the Appearance-level.

\subsection{Layout-Level Fine-Tuning}
\label{sec:layout-level-fine-tuning}

\medskip
\noindent\textbf{Layout Annotation.}\quad
The first step of the pipeline is to detect character (or grapheme cluster) locations. The character detector is a frozen pre-trained U-Net, CRAFT~\cite{baek2019character}. It produces the heatmap in~\cref{fig:human_in_loop_pipeline}(b).
Next, the 2D character locations we get from the heatmap in~\cref{fig:human_in_loop_pipeline}(b) are used by a Graph Neural Network based layout analysis backbone~\cite{chincholikar2026towards} to perform text-line segmentation. This graph-based problem formulation considers each character (or grapheme cluster) of the manuscript as a node, with edges connecting nodes of the same text-line to their neighbours in the same text-line. First, a preprocessing step is performed to get information-rich node and edge features using geometric inductive priors, after which the GNN performs binary edge classification to predict whether an edge exists between two nodes, giving us the predicted graph seen in~\cref{fig:human_in_loop_pipeline}(c). This problem formulation decomposes the text-line detection problem into two sub-problems: (i) character detection, and (ii) binary edge classification (connecting characters belonging to the same text-line together). This task decomposition enables the GNN to be pre-trained on \textit{large scale, diverse, synthetic layout data in the geometric domain} (only consisting of node locations and edge connections), while the character detection task is delegated to the pre-trained frozen CRAFT model~\cite{baek2019character}. (As illustrated in~\cref{fig:human_in_loop_pipeline} (b) to (c)).

% the layout analysis backbone has been shown~\cite{chincholikar2026towards} to be more data-efficient and stable over layout-level distribution shifts than competing open source methods SeamFormer~\cite{vadlamudi2023seamformer} and Doc-UFCN~\cite{Boillet_2021}.

\medskip
\noindent\textbf{Iterative GNN Fine-tuning.}\quad
The predictions of the GNN can be (optionally) corrected manually by the human user for two reasons: (a) to create a supervised training data point which is used to fine-tune the GNN, and (b) for manual correction at inference time. Using the annotated training data points of the target manuscript, we fine-tune a Graph Neural Network with the SplineCNN architecture~\cite{fey2018splinecnn}, which is pre-trained on diverse synthetic layouts in the geometric domain to perform a binary edge classification task, where an edge between two nodes is classified as \texttt{\textbf{1}} if the nodes are neighbours in the same text-line, and \texttt{\textbf{0}} otherwise. If doing manual correction at inference time, the user can also manually add/delete incorrect nodes (in addition to adding/deleting incorrect edges) to account for mistakes in character detection by CRAFT; however, these node manual corrections are not currently used to fine-tune the GNN, as it currently performs only a binary edge classification task. Text-Region annotations, as shown in~\cref{fig:page-10-annotation-views}(b) are also currently done manually.

\subsection{Appearance-Level Fine-Tuning}
\label{sec:appearance-level-fine-tuning}
Once the layout graph has been corrected, the pipeline prepares one text-line image per text-line, as shown in~\cref{fig:human_in_loop_pipeline}(d).

\medskip
\noindent\textbf{Curved Text Unwrapping.}\quad
In this step, if the detected text is curved, we unwrap it and convert the text from the graph-based format into a rectangular text-line image format, which the downstream text recognition OCR model (CNN--BiLSTM--CTC) requires. For each node along the text-line, we construct a local coordinate system: the tangent direction becomes the horizontal direction of the crop, and the normal direction becomes its vertical direction. Using information from the heatmap, we then sample the manuscript image in these local coordinates to form the rectangular image in~\cref{fig:human_in_loop_pipeline}(d). In this sense, we consider a circular text-line as a straight line locally; walking along its circular baseline unwraps it into a conventional left-to-right strip for OCR. This step is heuristic, as the circular text is cut at the topmost point in the global page coordinate system to define the start and end of the unwrapped line. It must also prevent diacritics from adjacent text-lines from entering the crop, and choose a reading direction.

\medskip
\noindent\textbf{OCR Fine-tuning.}\quad
The text recognition OCR model CNN--BiLSTM--CTC takes the (optionally unwrapped) and processed text-line image as input and predicts its text content as Unicode text, as shown in~\cref{fig:human_in_loop_pipeline}(e). The predictions can then be corrected by human experts to get ground-truth input-label pairs\textbf{(text-line image, corrected Unicode text)}. These input-label pairs are used to fine-tune the CNN--BiLSTM--CTC to adapt to the appearance-level distribution shift of the target manuscript, such as the scribe's writing style, period-specific writing conventions, page texture, and other nuisance factors. This adaption when done iteratively, allows the CNN--BiLSTM--CTC to make better predictions on subsequent pages and thus also reduces the burden of human annotation iteratively.

\subsection{Fine-tuning Configuration}
\label{sec:fine-tuning-configuration}
\noindent\textbf{System Requirements.}\quad
\label{sec:system_requirements}
All inference, annotation, and fine-tuning were done locally on a laptop with an NVIDIA GeForce RTX~4050 Laptop GPU, an AMD Ryzen processor, and 16\,GB RAM. The setup comprises a Vue.js front end and connects with the traditional pipeline Flask back end, which orchestrates the iterative fine-tuning and inference. After the user corrects a new page, the layout-level GNN and the
appearance-level CNN--BiLSTM--CTC text recognizer are both fine-tuned sequentially using
the newly corrected data. The resulting adapted checkpoints are then used to
predict the subsequent pages of the same manuscript. On average, adapting the annotation tool to each newly annotated page required 41.10\,s for GNN fine-tuning and 27.51\,s for CNN--BiLSTM--CTC fine-tuning.

\medskip
\noindent\textbf{Layout-level GNN fine-tuning.}\quad
We fine-tune the pre-trained GNN using the human-corrected layout graph of the target manuscript page. As the layout-level data is in the 2D geometric domain, where each page is represented by character-node locations and candidate edges rather than manuscript pixels, we can easily augment the corrected page layout 50 times using transformations such as warping, skewing, node drop-out, and node jitter~\cite{chincholikar2026towards}. We fine-tune all GNN parameters for 10 epochs
using Adam with a learning rate of $10^{-3}$ and a batch size of 4. To address the imbalance between positive and negative candidate edges, we use focal loss with $\alpha=0.9$ and $\gamma=2.0$. The learning rate is linearly warmed up during the first five epochs. Checkpoint selection maximises text-line F1 (IoU $\geq 0.5$) on the unaugmented fine-tuning pages themselves rather than a held-out split; the fold's test pages are reserved exclusively for the reported metrics.
% To reduce catastrophic forgetting, we also retain a deterministic 20\% replay sample of the augmented graphs from each previously corrected page; these earlier examples are included alongside the new page during fine-tuning.

\medskip
\noindent\textbf{Appearance-level text recognition OCR fine-tuning.}\quad
The pre-trained CNN--BiLSTM--CTC recognizer~\cite{chincholikar2025case}, which is based on the code provided by Clova AI’s deep-text-recognition-benchmark repository\footnote{\url{https://github.com/clovaai/deep-text-recognition-benchmark}} and the EasyOCR python package, is fine-tuned on the corrected input-label pairs \textbf{(text-line image, corrected Unicode text)} from the newly annotated page. We use
Adadelta with learning rate 0.2, $\rho=0.95$, and
$\epsilon=10^{-8}$ for 60 iterations, with batch size 1 and gradient clipping
at 5. Text-line images are resized to a height of 50 pixels and padded to the
maximum width within each batch. Checkpoint selection is in-sample: the accuracy and edit-distance snapshots are scored on unaugmented text-line image–label pairs from the training data, and the two are decided between by page CER on the fine-tuning pages; the fold's test pages are reserved for the reported metrics.
% The fine-tuning dataset also includes a
% random replay sample of 10 text-lines from previously corrected pages.

\medskip
\noindent\textbf{Annotation Cost.}\quad
The GNN and the OCR Recognition model are not supervised by the same annotations. The GNN learns from the corrected graph alone and never reads text, whereas the CNN--BiLSTM--CTC needs its input text-line images cut from an already corrected layout, which are then transcribed to get it's \textbf{(text-line image, corrected Unicode text)} input-label training pairs. Hence, the annotation cost of fine-tuning the OCR Recognition model on one page subsumes that of fine-tuning the GNN on one page.

\section{Experiment Setup}
\label{section: setup}

\medskip
\noindent\textbf{Data Splits.}\quad
For each manuscript, we create five folds using a fixed random seed. In each fold, three pages are reserved for fine-tuning the GNN and the CNN--BiLSTM--CTC, and all remaining pages form the held-out test set. We quantify the gains due to this fine-tuning on 1, 2, and 3 pages by using the held-out test data pages of the respective target manuscript, across 5 folds. Pages may reappear across folds, but the fine-tuning and test sets are disjoint within each fold. Each fine-tuned pipeline checkpoint is evaluated on the same test pages. 
% We also quantify the impact of manual layout-correction if done at inference time.

\medskip
\noindent\textbf{Multi-modal large language model pipeline.}\quad
We also evaluate four off-the-shelf multi-modal OCR systems: Gemini-3.5-Flash, OpenAI GPT-5.6-Terra, Claude-Sonnet-5, and Sarvam Vision Document Digitization. Gemini, OpenAI, and Claude receive the same resized manuscript-page image and the same end-to-end prompt (see Supplementary Material), which requests a diplomatic Unicode Devanagari transcription together with one polygon per visual text-line in a normalized $0$--$1000$ coordinate system following the conventions established in Pix2Seq~\cite{chen2021pix2seq} and PaLI~\cite{chen2023pali}, which may favour Gemini 3.5 Flash, as the respective research which set the convention, was done by researchers associated with Google. The JSON outputs are parsed into the standard PAGE-XML representation containing locations of the text-lines in a bounding polygon format, and the corresponding Unicode text content. As Sarvam uses its own Sanskrit document-digitization API and returns HTML without bounding polygons at the text-line level, we parse the HTML into the same PAGE-XML text-line representation, preserving their emitted order and Unicode text but leaving the geometric information of the text-line location empty. Due to this, Sarvam's Page-CER metric is evaluated by treating the emitted HTML text-line order as reading order. In the event that API failures or parsing failures occur, we retry 3 more times before considering the prediction of the page as a full failure. Multi-modal large language model predictions are evaluated for the exact same held-out test pages, across the exact five folds used for the traditional pipeline evaluation. The exact MMLLM model identifier requested and the HTTP endpoint it is requested from can be accessed in the Supplementary Material. The raw model responses are provided with the dataset released.
% We do not report the predictions of DeepSeek-OCR~\cite{wei2025deepseek} and Unlimited-OCR~\cite{yin2026unlimitedocrworks} as their predictions were unsatisfactory.

\medskip
\noindent\textbf{Metrics.}\quad
We quantify the transcription accuracy using standard metrics TextEdit~\cite{ouyang2025omnidocbenchbenchmarkingdiversepdf} and Page-CER~\cite{boillet2022robust,jimaging10030065,heidenreich2026gutenocr}. The TextEdit metric ignores line order and geometry, as it pairs each ground-truth line with at most one predicted line and counts unmatched lines as missing or extra text. To calculate TextEdit we use the exact same OmniDocBench
\texttt{simple\_match} score\footnote{Code (v1.5): \url{https://github.com/opendatalab/OmniDocBench/tree/v1_5},
pinned at commit \texttt{59b103c}.}. Each non-empty PAGE-XML \texttt{TextLine} is parsed as an atomic item using its direct \texttt{TextEquiv/Unicode} transcription and supplied as an individual text block to the matcher. \texttt{TextRegion} membership is ignored. To calculate Page,-CER we sort the ground-truth and predicted lines from top to bottom and left to right, join the text strings in that order, and then calculate the Character Edit Distance(CER).

\section{Results}

\begin{table*}[htbp]

\caption{
Benchmarking the performance of Multi-Modal Large Language Models using the metrics TextEdit and Page-CER. For Sarvam Vision, \textsuperscript{*} marks Page-CER computed by
treating the model's emitted HTML text-line order as reading order, as Sarvam does not provide text-line level bounding polygons. Each entry reports the metric value followed by
its 95\% confidence interval. Lower values are better.
}
\label{tab:e2e_models_by_manuscript}

\centering
\small
\setlength{\tabcolsep}{6pt}
\renewcommand{\arraystretch}{1.15}

\resizebox{\textwidth}{!}{%
\begin{tabular}{@{}llcc@{}}
\toprule

\textbf{Manuscript} &
\textbf{Multi-Modal LLM} &
\textbf{TextEdit $\downarrow$} &
\textbf{Page-CER $\downarrow$} \\

\midrule

% ---------------------------------------------------------------------------
% Moderate Layout / YAJN
% ---------------------------------------------------------------------------

\multirow{4}{*}{\texttt{Moderate Layout}} &
\texttt{Claude Sonnet 5} &
$0.60\,[0.45,\,0.76]$ &
$0.64\,[0.50,\,0.80]$ \\

&
\texttt{Gemini 3.5 Flash} &
\textbf{0.26\,[0.18,\,0.37]} &
\textbf{0.35\,[0.29,\,0.44]} \\

&
\texttt{GPT 5.6 Terra} &
$0.71\,[0.70,\,0.73]$ &
$0.74\,[0.73,\,0.76]$ \\

&
\texttt{Sarvam Vision} &
$0.33\,[0.31,\,0.36]$ &
$0.39\,[0.36,\,0.41]$\textsuperscript{*} \\

\midrule

% ---------------------------------------------------------------------------
% Dense Layout
% ---------------------------------------------------------------------------

\multirow{4}{*}{\texttt{Dense Layout}} &
\texttt{Claude Sonnet 5} &
$0.80\,[0.60,\,1.00]$ &
$0.85\,[0.65,\,1.00]$ \\

&
\texttt{Gemini 3.5 Flash} &
\textbf{0.30\,[0.27,\,0.34]} &
\textbf{0.44\,[0.38,\,0.47]} \\

&
\texttt{GPT 5.6 Terra} &
$0.79\,[0.77,\,0.80]$ &
$0.79\,[0.78,\,0.79]$ \\

&
\texttt{Sarvam Vision} &
$0.43\,[0.25,\,0.66]$ &
$0.64\,[0.47,\,0.79]$\textsuperscript{*} \\

\midrule

% ---------------------------------------------------------------------------
% Circular Layout / Circle-New
% ---------------------------------------------------------------------------

\multirow{4}{*}{\texttt{Circular Layout}} &
\texttt{Claude Sonnet 5} &
$0.94\,[0.83,\,1.00]$ &
$0.97\,[0.90,\,1.00]$ \\

&
\texttt{Gemini 3.5 Flash} &
\textbf{0.51\,[0.43,\,0.59]} &
\textbf{0.51\,[0.44,\,0.58]} \\

&
\texttt{GPT 5.6 Terra} &
$0.81\,[0.78,\,0.85]$ &
$0.79\,[0.76,\,0.82]$ \\

&
\texttt{Sarvam Vision} &
$0.63\,[0.49,\,0.77]$ &
$1.44\,[0.57,\,2.69]$\textsuperscript{*} \\

\bottomrule
\end{tabular}%
}

\vspace{1em}
\vspace{1em}

\caption{
Quantifying the gains due to fine-tuning the Traditional OCR pipeline across manuscripts using the metrics TextEdit and Page-CER. For each metric, \textbf{Iterative Fine-tuning (Fully Automatic)} runs the fine-tuned
pipeline inference fully automatically on the pages in the test data, whereas \textcolor{black}{\textbf{Iterative Fine-tuning (With Manual Layout Correction)}} denotes performing manual correction of predicted layouts before the downstream text recognition OCR. Metric values are followed by their 95\%
confidence intervals. Lower values are better.
% GT layout time is omitted as a separate column. For
% \texttt{Moderate Layout}, the GT layout time is
% $38.0\,[26.9,\,50.9]\,\mathrm{s}$. For \texttt{Dense Layout}, the GT layout
% time is $165.5\,[115.6,\,216.0]\,\mathrm{s}$. For
% \texttt{Circular Layout}, the GT layout time is
% $109.0\,[84.8,\,134.7]\,\mathrm{s}$.
}
\label{tab:traditional_pipeline}

\centering
\small
\setlength{\tabcolsep}{2.5pt}
\renewcommand{\arraystretch}{1.15}

\resizebox{\textwidth}{!}{%
\begin{tabular}{@{}lcc>{\color{black}}c@{\hspace{5pt}}c>{\color{black}}c@{}}
\toprule

\multirow{2}{*}{\textbf{Manuscript}} &
\multirow{2}{*}{\shortstack{\textbf{Pages}\\\textbf{Fine-tuned}}} &
\multicolumn{2}{c}{\textbf{TextEdit $\downarrow$}} &
\multicolumn{2}{c}{\textbf{Page-CER $\downarrow$}} \\

\cmidrule(lr){3-4}
\cmidrule(lr){5-6}

& &
\shortstack[t]{\textbf{Iterative}\\\textbf{Fine-tuning}\\\textbf{(Fully}\\\textbf{Automatic)}} &
\shortstack[t]{\textbf{Iterative}\\\textbf{Fine-tuning}\\{\color{black}\textbf{(With Manual}}\\{\color{black}\textbf{Layout Correction)}}} &
\shortstack[t]{\textbf{Iterative}\\\textbf{Fine-tuning}\\\textbf{(Fully}\\\textbf{Automatic)}} &
\shortstack[t]{\textbf{Iterative}\\\textbf{Fine-tuning}\\{\color{black}\textbf{(With Manual}}\\{\color{black}\textbf{Layout Correction)}}} \\

\midrule

% ---------------------------------------------------------------------------
% Moderate Layout
% ---------------------------------------------------------------------------

\multirow{4}{*}{\texttt{Moderate Layout}} &
0 &
$0.34\,[0.32,\,0.38]$ &
$0.30\,[0.28,\,0.32]$ &
$0.32\,[0.30,\,0.36]$ &
$0.30\,[0.28,\,0.32]$ \\

&
1 &
$0.28\,[0.26,\,0.30]$ &
$0.24\,[0.22,\,0.25]$ &
$0.25\,[0.24,\,0.26]$ &
$0.23\,[0.22,\,0.24]$ \\

&
2 &
$0.24\,[0.22,\,0.26]$ &
$0.20\,[0.19,\,0.21]$ &
$0.22\,[0.20,\,0.23]$ &
$0.20\,[0.19,\,0.21]$ \\

&
3 &
\textbf{0.23\,[0.21,\,0.24]} &
$0.19\,[0.17,\,0.20]$ &
\textbf{0.20\,[0.19,\,0.22]} &
$0.18\,[0.17,\,0.19]$ \\

\midrule

% ---------------------------------------------------------------------------
% Dense Layout
% ---------------------------------------------------------------------------

\multirow{4}{*}{\texttt{Dense Layout}} &
0 &
$0.33\,[0.29,\,0.38]$ &
$0.21\,[0.20,\,0.23]$ &
$0.34\,[0.30,\,0.36]$ &
$0.25\,[0.23,\,0.27]$ \\

&
1 &
$0.32\,[0.27,\,0.37]$ &
$0.19\,[0.19,\,0.21]$ &
$0.33\,[0.30,\,0.36]$ &
$0.23\,[0.22,\,0.24]$ \\

&
2 &
$0.28\,[0.24,\,0.34]$ &
$0.18\,[0.17,\,0.19]$ &
$0.29\,[0.27,\,0.30]$ &
$0.22\,[0.21,\,0.23]$ \\

&
3 &
\textbf{0.26\,[0.21,\,0.32]} &
$0.16\,[0.16,\,0.17]$ &
\textbf{0.26\,[0.25,\,0.27]} &
$0.20\,[0.19,\,0.21]$ \\

\midrule

% ---------------------------------------------------------------------------
% Circular Layout
% ---------------------------------------------------------------------------

\multirow{4}{*}{\texttt{Circular Layout}} &
0 &
$0.48\,[0.44,\,0.52]$ &
$0.35\,[0.33,\,0.38]$ &
$0.52\,[0.48,\,0.56]$ &
$0.40\,[0.37,\,0.44]$ \\

&
1 &
$0.46\,[0.40,\,0.51]$ &
$0.33\,[0.30,\,0.36]$ &
$0.51\,[0.47,\,0.55]$ &
$0.38\,[0.35,\,0.42]$ \\

&
2 &
$0.44\,[0.39,\,0.49]$ &
$0.32\,[0.29,\,0.35]$ &
$0.50\,[0.45,\,0.54]$ &
$0.37\,[0.33,\,0.40]$ \\

&
3 &
\textbf{0.40\,[0.35,\,0.45]} &
$0.30\,[0.27,\,0.33]$ &
\textbf{0.48\,[0.43,\,0.53]} &
$0.36\,[0.33,\,0.40]$ \\

\bottomrule
\end{tabular}%
}

\end{table*}

\medskip
\noindent\textbf{Off-the-Shelf Multi-modal LLMs.}\quad
~\cref{tab:e2e_models_by_manuscript} reports the end-to-end performance of the four off-the-shelf multi-modal LLMs on the held-out test pages, aggregated over five folds as described in~\cref{section: setup}. Gemini 3.5 Flash performs best on every manuscript and on both metrics, with Sarvam Vision being the second in every case. Every multi-modal LLM ranks the difficulty of the three manuscripts in the same order: Moderate $<$ Dense $<$ Circular on both metrics. The Page-CER of Sarvam Vision on the Circular Layout Manuscript reaches $1.44\,[0.57,\,2.69]$, which is caused due to "runaway generation" type of hallucinations in which the model over-generates random tokens, and thus causes the character edit distance to exceed the number of ground-truth characters.

\medskip
\noindent\textbf{Off-the-Shelf Traditional Pipeline.}\quad
The traditional pipeline, when used off-the-shelf (with no fine-tuning on the target manuscript and no manual layout correction), performs comparably with Gemini 3.5 Flash as seen the 0-page fine-tuned rows of the "Iterative Fine-tuning (Fully Automatic)" subcolumns of~\cref{tab:traditional_pipeline}, and the Gemini 3.5 Flash rows in~\cref{tab:e2e_models_by_manuscript}. For the PAGE-CER metric, the off-the-shelf traditional pipeline outperforms Gemini 3.5 Flash on Moderate and Dense Layout Manuscripts but underperforms on the Circular Layout Manuscript. For the TextEdit metric, the traditional pipeline performs worse than Gemini on Moderate and Dense Layout Manuscripts but better on the Circular Layout Manuscript.

% In this setting, TextEdit and Page-CER of the off-the-shelf traditional pipeline are $0.34$ and $0.32$ on the Moderate Layout Manuscript, $0.33$ and $0.34$ on the Dense Layout Manuscript, and $0.48$ and $0.52$ on the Circular Layout Manuscript. 

\medskip
\noindent\textbf{Iteratively Fine-tuned Traditional Pipeline (Fully Automatic).}\quad
While off-the-shelf results are important when performing OCR in bulk quantities, the benefits of iterative fine-tuning become more apparent when performing Historical OCR, where the traditional pipeline needs to adapt to target manuscript heterogeneity. In this setting, fine-tuning the GNN and the CNN--BiLSTM--CTC of the traditional pipeline on pages of the target manuscript helps it adapt to the distribution of the target manuscript at the layout-level and appearance-level, respectively, and improves prediction quality monotonically with each new page fine-tuned, as seen in "Iterative Fine-tuning  (Fully Automatic)" subcolumns of~\cref{tab:traditional_pipeline}. Fine-tuning both models on three pages reduces TextEdit by $34.1\%$ on the Moderate Layout Manuscript, $22.9\%$ on the Dense Layout Manuscript, and $15.5\%$ on the Circular Layout Manuscript. The corresponding Page-CER reductions are $37.5\%$, $22.9\%$, and $8.1\%$. Once fine-tuned on 3 pages of each target manuscript, the traditional pipeline performs substantially better than Gemini 3.5 Flash on all three manuscripts on both metrics. 

% The text-line segmentation errors from the GNN at inference time can mask the benefits of fine-tuning the CNN--BiLSTM--CTC model, since even a strong text recognition OCR model cannot reliably process incorrectly segmented text-line images due to incorrect splitting or merging of adjacent text-lines at the layout-level.

\medskip
\noindent\textbf{Iteratively Fine-tuned Traditional Pipeline (with Manual Layout Correction).}\quad
In the above Iterative Fine-tuning (Fully Automatic) setting, the fine-tuned pipeline does inference fully automatically, without any manual layout correction. However, enabling historians and scholars to perform Manual layout correction at inference time (on the test set in this experiment) can ensure that there are no costly layout analysis mistakes that can cause disastrous consequences for the downstream line level text recognition OCR task~\cite{a16030136}. In other words, for the "Iterative Fine-tuning (Fully Automatic)" subcolumn, both the GNN and the CNN--BiLSTM--CTC perform inference automatically when fine-tuned on up to three pages. In comparison, in the "Iterative Fine-tuning (With Manual Layout Correction)" setting, the GNN's predictions are manually corrected at inference time, providing error-free text-line detection, on which the fine-tuned CNN--BiLSTM--CTC performs inference automatically. The performance gap between the columns "Iterative Fine-tuning (Fully Automatic)" and "Iterative Fine-tuning (With Manual Layout Correction)" quantifies the effects of object-level layout-detection errors—and, equivalently, quantifies the benefits of correcting them. We observe that the benefit of manual layout correction is minimal for Moderate Layout Manuscript ($12.6\%$ TextEdit and $17.8\%$ Page-CER), and is the greatest for Dense Layout Manuscript ($36.2\%$ TextEdit and $36.9\%$ Page-CER). Across all three manuscripts, the traditional pipeline achieves the best performance when fine-tuned on three pages with manual layout corrections enabled. In this experiment, $38.0$, $165.5$, and $109.0$ seconds per page were required on average to perform manual layout correction on the Moderate, Dense, and Circular Layout Manuscripts, respectively. 

% These seconds are already contained in the cost of every page used to fine-tune the CNN--BiLSTM--CTC (in terms of annotation time), whose text-line images are cut from a corrected layout, so layout correction is not an additional expense for an annotator transcribing the page in any case, and the corrected graph it yields supervises the GNN for free. The converse does not hold, since a corrected graph requires no transcription: Read-Mode effort is not instrumented, but at a conservative estimate of one hour per page, a page annotated for OCR fine-tuning costs $96\times$, $23\times$, and $34\times$ one annotated for GNN fine-tuning, respectively.
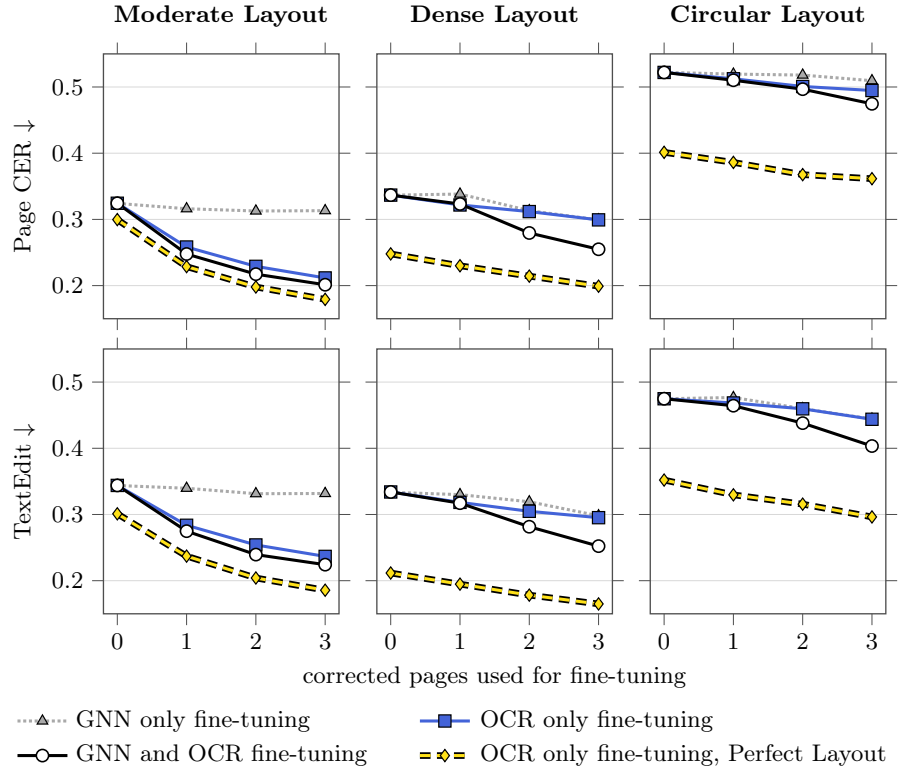
\begin{figure*}[!t]
\centering

% ---- colour-blind-safe palette ---------------------------------------------
\definecolor{cbYellow}{HTML}{FFE119}
\definecolor{cbBlue}{HTML}{4363D8}
\definecolor{cbGrey}{HTML}{A9A9A9}
\definecolor{cbWhite}{HTML}{FFFFFF}
\definecolor{cbBlack}{HTML}{000000}

% ---- series styles ---------------------------------------------------------
% Distinguished by colour, dash pattern and marker at once, so the panels stay
% readable in greyscale and under any colour-vision deficiency. Defined with
% \tikzset (not \pgfplotsset) so the same styles drive the plots and the
% hand-built legend keys below.
\tikzset{
  drBase/.style   = {line width=1.15pt, mark size=2.3pt},
  drGnn/.style    = {drBase, color=cbGrey, densely dotted, mark=triangle*,
                     mark options={solid, draw=cbBlack, fill=cbGrey, line width=0.45pt}},
  drOcr/.style    = {drBase, color=cbBlue, mark=square*,
                     mark options={solid, draw=cbBlack, fill=cbBlue, line width=0.45pt}},
  drJoint/.style  = {drBase, color=cbBlack, mark=*,
                     mark options={solid, draw=cbBlack, fill=cbWhite, line width=0.6pt}},
  drGt/.style     = {drBase, color=cbYellow, dash pattern=on 4pt off 2.4pt, mark=diamond*,
                     mark options={solid, draw=cbBlack, fill=cbYellow, line width=0.45pt}},
  % black outline under the yellow dashes; yellow alone is illegible on white
  drGtHalo/.style = {color=cbBlack, line width=2.5pt, mark=none,
                     dash pattern=on 4pt off 2.4pt},
}

\begin{tikzpicture}
\begin{groupplot}[
  group style={
    group size=3 by 2,
    horizontal sep=0.5cm,
    vertical sep=0.4cm,
    y descriptions at=edge left,
    x descriptions at=edge bottom,
  },
  % panel width follows \textwidth, so the figure fits any two-column template;
  % the subtracted 0.95cm pays for the shared y label and tick labels
  width={\dimexpr0.333\textwidth-0.95cm\relax}, height=3.5cm,
  scale only axis,
  xmin=-0.2, xmax=3.2, xtick={0,1,2,3},
  ymin=0.15, ymax=0.55, ytick={0.20,0.30,0.40,0.50},
  yticklabel style={/pgf/number format/fixed, /pgf/number format/precision=2},
  ymajorgrids, grid style={draw=cbGrey!45, line width=0.3pt},
  axis line style={draw=cbBlack!70, line width=0.5pt},
  tick align=outside, tick style={draw=cbBlack!70},
  title style={align=center, yshift=1pt},
  label style={font=\small}, tick label style={font=\small},
  clip=false,
]
% ---- row 1: Page CER ----------------------------------------------
\nextgroupplot[
  title={\textbf{Moderate Layout}},
  ylabel={Page CER $\downarrow$}]
\addplot[drGtHalo, forget plot] coordinates {(0,0.2997) (1,0.2287) (2,0.1979) (3,0.1792)};
\addplot[drGnn] coordinates {(0,0.3242) (1,0.3161) (2,0.3127) (3,0.3132)};
\addplot[drOcr] coordinates {(0,0.3244) (1,0.2583) (2,0.2291) (3,0.2116)};
\addplot[drJoint] coordinates {(0,0.3244) (1,0.2477) (2,0.2173) (3,0.2013)};
\addplot[drGt] coordinates {(0,0.2997) (1,0.2287) (2,0.1979) (3,0.1792)};
\nextgroupplot[
  title={\textbf{Dense Layout}}]
\addplot[drGtHalo, forget plot] coordinates {(0,0.2479) (1,0.2300) (2,0.2143) (3,0.1991)};
\addplot[drGnn] coordinates {(0,0.3366) (1,0.3383) (2,0.3132) (3,0.2993)};
\addplot[drOcr] coordinates {(0,0.3367) (1,0.3218) (2,0.3118) (3,0.2993)};
\addplot[drJoint] coordinates {(0,0.3367) (1,0.3233) (2,0.2795) (3,0.2549)};
\addplot[drGt] coordinates {(0,0.2479) (1,0.2300) (2,0.2143) (3,0.1991)};
\nextgroupplot[
  title={\textbf{Circular Layout}}]
\addplot[drGtHalo, forget plot] coordinates {(0,0.4013) (1,0.3861) (2,0.3676) (3,0.3615)};
\addplot[drGnn] coordinates {(0,0.5221) (1,0.5195) (2,0.5180) (3,0.5095)};
\addplot[drOcr] coordinates {(0,0.5220) (1,0.5124) (2,0.5008) (3,0.4947)};
\addplot[drJoint] coordinates {(0,0.5220) (1,0.5102) (2,0.4967) (3,0.4746)};
\addplot[drGt] coordinates {(0,0.4013) (1,0.3861) (2,0.3676) (3,0.3615)};
% ---- row 2: TextEdit ----------------------------------------------
\nextgroupplot[
  ylabel={TextEdit $\downarrow$}]
\addplot[drGtHalo, forget plot] coordinates {(0,0.3010) (1,0.2371) (2,0.2041) (3,0.1856)};
\addplot[drGnn] coordinates {(0,0.3438) (1,0.3396) (2,0.3314) (3,0.3317)};
\addplot[drOcr] coordinates {(0,0.3439) (1,0.2837) (2,0.2540) (3,0.2367)};
\addplot[drJoint] coordinates {(0,0.3439) (1,0.2748) (2,0.2394) (3,0.2242)};
\addplot[drGt] coordinates {(0,0.3010) (1,0.2371) (2,0.2041) (3,0.1856)};
\nextgroupplot[
  xlabel={corrected pages used for fine-tuning}]
\addplot[drGtHalo, forget plot] coordinates {(0,0.2116) (1,0.1948) (2,0.1782) (3,0.1649)};
\addplot[drGnn] coordinates {(0,0.3335) (1,0.3300) (2,0.3192) (3,0.2981)};
\addplot[drOcr] coordinates {(0,0.3340) (1,0.3183) (2,0.3048) (3,0.2952)};
\addplot[drJoint] coordinates {(0,0.3340) (1,0.3171) (2,0.2815) (3,0.2522)};
\addplot[drGt] coordinates {(0,0.2116) (1,0.1948) (2,0.1782) (3,0.1649)};
\nextgroupplot
\addplot[drGtHalo, forget plot] coordinates {(0,0.3520) (1,0.3297) (2,0.3156) (3,0.2963)};
\addplot[drGnn] coordinates {(0,0.4748) (1,0.4767) (2,0.4599) (3,0.4442)};
\addplot[drOcr] coordinates {(0,0.4748) (1,0.4684) (2,0.4596) (3,0.4438)};
\addplot[drJoint] coordinates {(0,0.4748) (1,0.4643) (2,0.4380) (3,0.4034)};
\addplot[drGt] coordinates {(0,0.3520) (1,0.3297) (2,0.3156) (3,0.2963)};
\end{groupplot}
\end{tikzpicture}

\smallskip

{\small
\begin{tabular}{@{}c@{\;}l@{\qquad}c@{\;}l@{}}
  \tikz[baseline=-0.55ex]{%
    \draw[drGnn, mark=none] (0,0) -- (0.66,0);
    \draw[drGnn, only marks] plot coordinates {(0.33,0)};} & GNN only fine-tuning &
  \tikz[baseline=-0.55ex]{%
    \draw[drOcr, mark=none] (0,0) -- (0.66,0);
    \draw[drOcr, only marks] plot coordinates {(0.33,0)};} & OCR only fine-tuning \\[2pt]
  \tikz[baseline=-0.55ex]{%
    \draw[drJoint, mark=none] (0,0) -- (0.66,0);
    \draw[drJoint, only marks] plot coordinates {(0.33,0)};} & GNN and OCR fine-tuning &
  \tikz[baseline=-0.55ex]{%
    \draw[drGtHalo] (0,0) -- (0.66,0);
    \draw[drGt, mark=none] (0,0) -- (0.66,0);
    \draw[drGt, only marks] plot coordinates {(0.33,0)};} & OCR only fine-tuning, Perfect Layout
\end{tabular}}

\caption{
We fine-tune the traditional pipeline up to three pages (across the same 5-fold train-test split as described in~\cref{section: setup}) using three ablations: one where only the GNN is fine-tuned at layout-level, one where only the CNN-BiLSTM-CTC OCR recognition model is fine-tuned at the appearance-level, and one where both are fine-tuned. We also report a fourth ablation where a human annotator manually corrects the layout of the test set pages at inference time, and an iteratively fine-tuned CNN-BiLSTM-CTC OCR recognition model is used to predict the text content from the text-line images. This fourth ablation is meant to mimic the expected working conditions of the Traditional Pipeline Annotation Tool, where manual layout correction takes a few minutes at most.}
\label{fig:diminishing_returns_combined}

\end{figure*}
\medskip
\noindent\textbf{Fine-tuning Ablations.}\quad
~\cref{fig:diminishing_returns_combined} compares the gains in downstream OCR accuracy due to GNN-only fine-tuning, OCR-only fine-tuning, and combined fine-tuning. For the Moderate Layout manuscript, we observe that GNN-only layout-level fine-tuning does not help as much as the OCR-only fine-tuning because the pre-trained GNN's layout predictions are already close to ground-truth, whereas fine-tuning the OCR Recognition model displays rapid adaptation to the target manuscript at the appearance-level. For Dense Layout and Circular Layout manuscripts, GNN-only and OCR-only gains due to fine-tuning are comparable on the TextEdit metric. The combined fine-tuning of GNN and OCR gives compounded improvement. On Circular, fine-tuning both models beats the sum of the two single-model gains by 19\% (TextEdit). Considering the annotation cost in terms of annotation time, this compounding is free: a page supervising the OCR recognition model must have its layout corrected before its text-line images can be cut, so the GNN and OCR combined fine-tuning (black, open circles) costs no more than OCR-only fine-tuning (blue, filled squares), while GNN-only fine-tuning (grey dotted, triangles) requires manual layout correction alone.

\section{Discussion}
When used without adaptation to the target manuscript, the traditional pipeline achieves comparable accuracy with the best-performing Multi-Modal LLM Gemini 3.5 Flash.

% (~\cref{tab:e2e_models_by_manuscript} and~\cref{tab:traditional_pipeline}).
However, fine-tuning the GNN and the CNN--BiLSTM--CTC of the Traditional pipeline on three corrected pages reduces TextEdit by up to $34.1\%$ and Page-CER by up to $37.5\%$, thus achieving better results than Gemini 3.5 Flash on all three manuscripts on both metrics. Fine-tuning on each new page thus effectively reduces the human annotation effort required for the next page. This is desirable especially when the data is scarce, and annotation is costly and time-consuming.

In addition to fine-tuning, performing manual layout correction at inference time removes the pipeline's object-level layout-detection errors, such as incorrectly merged or split text-lines by the GNN, or incorrectly predicted missing or extra nodes by CRAFT. The errors that remain are attributable to the text-line unwrapping and processing, and the fine-tuned iterations of the text-line image recognition model CNN--BiLSTM--CTC. When combined with iterative fine-tuning, inference time Manual layout correction further reduced the TextEdit by up to $51.3\%$, and roughly required tens of seconds to a few minutes per page.

We thus conclude that the specialized Traditional pipeline, which leverages domain knowledge in various ways, is well suited to digitize Sanskrit manuscripts where the data is scarce, and annotation is time-consuming and expensive. However, a limitation of the pipeline is that it is brittle~\cite{heidenreich2026gutenocr}, because of it's step by step nature, use of heuristics in unwrapping and processing the text-line images, and it's dependence on the frozen pre-trained character detector CRAFT. 

We use this brittle but task-specific and locally fine-tunable traditional pipeline to bootstrap the creation of a richly annotated dataset containing layout-level annotations and downstream Unicode transcriptions, represented in both a graph-based format and the standard PAGE-XML format. Notably, the final outputs of the digitization task—namely, text-line locations and Unicode text content—can be \textit{externally verified}, making them suitable for post-training and fine-tuning modern multimodal large language models. This direction is promising because multimodal large language models are less susceptible to the brittleness of task-specific traditional pipelines. However, they are more data-intensive and require an initial curated dataset, which the proposed traditional pipeline can effectively bootstrap.

\section*{Acknowledgements}
The authors wish to express their thanks to Lalchand Research Library, DAV College, Chandigarh, India, the eGangotri Project, and the Gyan Bharatam Project, for making manuscript data publicly available for educational and research purposes.
The authors also wish to express their gratitude to the anonymous reviewers, Dr. Petar Veličković, Dr. Dhaval Patel, Dr. Oliver Hellwig and Dr. Tarinee Awasthi for their invaluable support and feedback. The authors also wish to thank their colleagues Shagun Dwivedi, Janhavi Vaishampayan and Ansh Kushwaha for reviewing this work, and for their insightful suggestions.

% BibTeX users should specify bibliography style 'splncs04'.
% References will then be sorted and formatted in the correct style.
\bibliographystyle{splncs04}
\bibliography{gram2026}

\clearpage
\section{Supplementary Material}
\subsection{Transcription using Off-the-shelf Multi-Modal LLMs}
Gemini-3.5-Flash, OpenAI GPT-5.6-Terra, Claude-Sonnet-5 received the exact same manuscript-page image and the same end-to-end prompt shown in~\cref{gemini_end_to_end} below, which requests a diplomatic Unicode Devanagari transcription together with one polygon per visual text-line in a normalized $0$--$1000$ coordinate system following the conventions established in Pix2Seq~\cite{chen2021pix2seq} and PaLI~\cite{chen2023pali}. This convention might favour Gemini 3.5 Flash, as the research papers which set the convention were authored by researchers associated with Google.
\label{gemini_end_to_end}
\begin{lstlisting}
prompt_text = """
You are an expert Indologist and Paleographer specializing in handwritten Sanskrit manuscripts.
Your Task: Perform a diplomatic transcription (OCR) of the manuscript image and provide text-line geometry.
CRITICAL INSTRUCTIONS:
1. Output Format: Output ONLY raw valid JSON. No Markdown.
2. Coordinates: Coordinates are normalized from 0 to 1000, where [0,0] is top-left and [1000,1000] is bottom-right.
3. Geometry: For every visual text-line, output polygon_2d as [[y,x], ...]. Use a tight polygon following the visible line. If the line is straight and rectangular, box_2d [ymin,xmin,ymax,xmax] is also acceptable. For curved or circular lines, polygon_2d is mandatory.
4. Granularity: Transcribe at the visual text-line level.
5. Script: Unicode Devanagari.
JSON SCHEMA:
{
  "status": "success",
  "regions": [
    {
      "id": "region_0",
      "type": "main_text",
      "polygon_2d": [[y,x], [y,x], [y,x]],
      "box_2d": [ymin, xmin, ymax, xmax],
      "lines": [
        {
          "id": "line_0",
          "polygon_2d": [[y,x], [y,x], [y,x]],
          "box_2d": [ymin, xmin, ymax, xmax],
          "text": "Transcribed text here"
        }
      ]
    }
  ]
}
"""
\end{lstlisting}

\subsection{Multi-Modal LLM Provider Information}
The exact MMLLM model identifier requested, and the
HTTP endpoint it is requested from is shown in~\cref{tab:vlm_providers}. The exact model responses are documented in the dataset released with this paper.

\begin{table*}[t]

\caption{
The four end-to-end VLM baselines: the exact model identifier requested and the
HTTP endpoint it is requested from. Gemini, OpenAI, and Claude receive the resized page image followed by the same
\texttt{VLM\_END\_TO\_END\_PROMPT} and return JSON; Sarvam receives the image
alone and returns HTML, hence its different call shape.
}
\label{tab:vlm_providers}

\centering
\small
\setlength{\tabcolsep}{5pt}
\renewcommand{\arraystretch}{1.2}

\begin{tabular}{@{}
  >{\raggedright\arraybackslash}p{0.13\textwidth}
  >{\raggedright\arraybackslash}p{0.19\textwidth}
  @{\hspace{22pt}}
  >{\raggedright\arraybackslash}p{0.60\textwidth}@{}}

\toprule

\textbf{Method} &
\textbf{Model} &
\textbf{API endpoint} \\

\midrule

\texttt{gemini\_e2e} &
\texttt{gemini-3.5-flash} &
\footnotesize\ttfamily POST \url{https://generativelanguage.googleapis.com/v1beta/models/gemini-3.5-flash:generateContent} \\

\addlinespace

\texttt{openai\_e2e} &
\texttt{gpt-5.6-terra} &
\footnotesize\ttfamily POST \url{https://api.openai.com/v1/responses} \\

\addlinespace

\texttt{claude\_e2e} &
\texttt{claude-sonnet-5} &
\footnotesize\ttfamily POST \url{https://api.anthropic.com/v1/messages} \\

\addlinespace

\texttt{sarvam\_e2e} &
\texttt{sarvam-vision}\textsuperscript{$\dagger$} &
\footnotesize\ttfamily
POST \url{https://api.sarvam.ai/doc-digitization/job/v1}
\rmfamily\textsuperscript{$\ddagger$}\ttfamily\newline
\rmfamily and, relative to that base:\ttfamily\newline
POST \url{/upload-files}\newline
POST \url{/{job_id}/start}\newline
GET\phantom{T} \url{/{job_id}/status}\newline
POST \url{/{job_id}/download-files} \\

\bottomrule
\end{tabular}

\vspace{5pt}

\begin{minipage}{\textwidth}
\footnotesize\raggedright
\textsuperscript{$\dagger$}~Sarvam Vision Document Digitization API accepts no model
parameter: the request body carries only \texttt{language=sa-IN} and
\texttt{output\_format=html}. \texttt{sarvam-vision} is the identifier recorded
in \texttt{VLM\_PROVIDER\_SPECS} and pinned into the cache fingerprint, not a
value sent over the wire.

\smallskip
\textsuperscript{$\ddagger$}~Sarvam Vision is asynchronous and job-based: one page is
one job, so a single page costs five calls in the order shown (create, upload,
start, poll status until terminal, download). \texttt{upload-files} and
\texttt{download-files} return presigned object-storage URLs; the page image and
the result ZIP transfer over those, not over \texttt{api.sarvam.ai}.
\end{minipage}

\end{table*}

\end{document}